\documentclass[lettersize,journal]{IEEEtran}
\usepackage{amsmath,amsfonts}
\usepackage{algorithmicx}
\usepackage{algorithm}
\usepackage{algpseudocode}
\usepackage{array}
\usepackage[caption=false,font=normalsize,labelfont=sf,textfont=sf]{subfig}
\usepackage{textcomp}
\usepackage{stfloats}
\usepackage{url}
\usepackage{verbatim}
\usepackage{graphicx}
\usepackage{cite}
\usepackage{color}
\usepackage{booktabs}
\usepackage{colortbl}
\usepackage{soul}
\definecolor{grey}{RGB}{160,160,160}
\definecolor{lightgray}{RGB}{235,235,235}
\usepackage{doi}
\usepackage{bbding}
\usepackage{pifont}
\usepackage{wasysym}
\usepackage{amssymb}
\usepackage{subcaption}
\usepackage{subfloat}
\usepackage{multirow}
\usepackage{makecell}
\usepackage{svg}
\usepackage{alltt}
\usepackage{multirow}

\usepackage{caption}
\usepackage{wrapfig}
\begin{document}

\title{InPK: Infusing Prior Knowledge into Prompt for Vision-Language Models}

\author{Shuchang Zhou, Jiwei Wei$^{\ast}$, Shiyuan He, Yuyang Zhou, Chaoning Zhang, \\Jie Zou, Ning Xie, Yang Yang,~\IEEEmembership{Senior Member,~IEEE}
\thanks{Shuchang Zhou, Shiyuan He, Chaoning Zhang, Jie Zou, and Ning Xie are with the Center for Future Media and the School of Computer Science and Engineering, University of Electronic Science and Technology of China, Chengdu 611731, China.}
\thanks{Yuyang Zhou is with Hainan University, Haikou, China.}
\thanks{Yang Yang, and Jiwei Wei are with the Center for Future Media and School of Computer Science and Engineering, University of Electronic Science and Technology of China, Chengdu 611731, China, and also with the Institute of Electronic and Information Engineering, University of Electronic Science and Technology of China, Dongguan 523808, China.}
\thanks{Corresponding author: Jiwei Wei. Email: mathematic6@gmail.com.}
}



\maketitle

\begin{abstract}
Prompt tuning has emerged as a prevalent approach for adapting VLMs to downstream visual recognition tasks in zero/few-shot settings.  While some prompting techniques leverage prior knowledge for its richness, the random initialization and disconnection of learnable tokens from this knowledge often lead to overfitting to seen classes and poor generalization across unseen domains. To address this issue, we propose the InPK model, which infuses class-specific prior knowledge into the learnable tokens during initialization, thus enabling the model to explicitly focus on class-relevant information. Furthermore, to mitigate the weakening of class information by multi-layer encoders, we continuously reinforce the interaction between learnable tokens and prior knowledge across multiple feature levels. This progressive interaction allows the learnable tokens to better capture the fine-grained differences and universal visual concepts within prior knowledge, enabling the model to extract more discriminative and generalized features. Even for unseen classes, the learned interaction allows the model to capture their common representations and infer their appropriate positions within the existing semantic structure. Moreover, we introduce a learnable text-to-vision projection layer to accommodate the text adjustments, ensuring better alignment of visual-text semantics. Extensive experiments on 11 recognition datasets show that InPK significantly outperforms state-of-the-art methods in multiple zero/few-shot image classification tasks.
\end{abstract}

\begin{IEEEkeywords}
Prompt tuning, vision-language model, few-shot classification.
\end{IEEEkeywords}

\section{Introduction}
\label{sec:intro}
\IEEEPARstart{I}{n} recent years, Vision-Language Models (VLMs) pre-trained on large-scale image-text datasets, such as CLIP~\cite{radford2021learning}, have demonstrated strong generalization capabilities across downstream tasks. Typically, VLMs employ contrastive loss to optimize both image and text encoders, aligning features from both modalities within a shared embedding space.  To further enhance model adaptation to specific tasks or datasets, particularly in zero/few-shot settings, the community has proposed numerous solutions~\cite{gao2024clip,zhang2022tip,zhou2022learning,zhou2022conditional,khattak2023maple,khattak2023self}. Among these, prompt tuning~\cite{zhou2022learning,zhou2022conditional,khattak2023maple,khattak2023self} has emerged as a popular strategy for adapting VLMs to various visual recognition tasks in zero/few-shot scenarios.
\begin{figure*}[t]
  \centering
     \scalebox{1}{
        \includegraphics[width=1\linewidth]{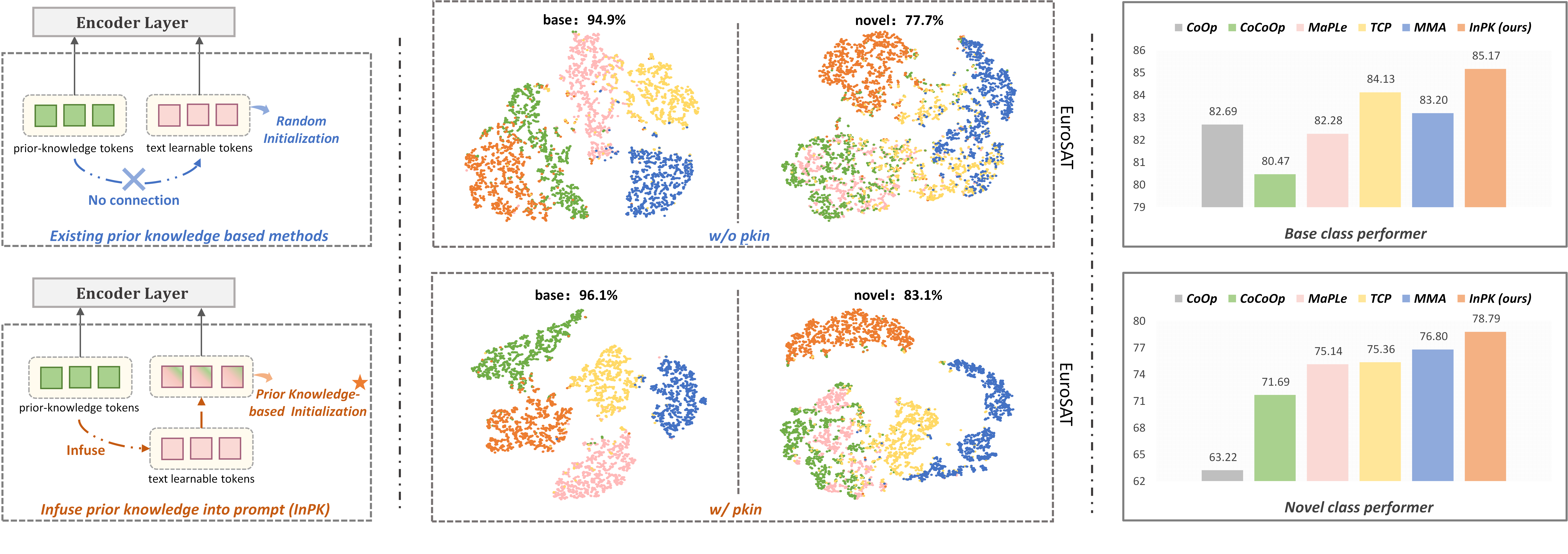} 
     }
    \vspace{-15pt}
   \caption{\textbf{(a)} Existing prior knowledge-based prompting techniques initialize learnable tokens randomly, resulting in task-agnostic token representations that prolong the optimization trajectory and increase susceptibility to local optima (illustrated in \textcolor[RGB]{64,114,196}{top}). Our approach infuses prior knowledge into learnable tokens before feeding them into the encoder layers each time, providing a prior knowledge-based initialization that explicitly emphasizes class-relevant information (illustrated in \textcolor[RGB]{197,90,17}{bottom}). \textbf{(b)} t-SNE visualization of the feature manifolds for our method with and without prior knowledge infusion (pkin) on the EuroSAT dataset. Our method with pkin shows tighter intra-class distances, indicating enhanced consistency, and larger inter-class distances,n reflecting improved class discrimination. \textbf{(c)} In the base-to-base/base-to-novel setting, our method significantly outperforms the state-of-the-art methods regarding average results on 11 recognition datasets.}
    \vspace{-10pt}
   \label{fig:knowledge_method}
\end{figure*}
 
Prompt tuning techniques have gradually gained favor in adapting Vision and Language Models (VLMs) to downstream tasks without the need to fully retrain the original models. For example, CoOp~\cite{zhou2022learning} and CoCoOp~\cite{zhou2022conditional} replace fixed prompts with learnable tokens, achieving substantial improvements through fine-tuning on a limited number of samples. Recent works~\cite{jia2022visual,khattak2023self,khattak2023maple,xing2023dual} extend this concept to multimodal prompt tuning, fine-tuning both text and visual branches to improve performance on both base (seen) and novel (unseen) classes. However, since learnable tokens are primarily optimized for base classes during training, prompt-based models often overfit to task-specific data distributions. 

Recently, a series of studies~\cite{lee2024coapt,menon2022visual,wang2024learning,kan2023knowledge,royconsistency} introduce prior knowledge to enhance object understanding and build more coherent text-visual mappings in the semantic space. While these methods mitigate overfitting to some extent, the random initialization of learnable tokens and their disconnection from prior knowledge (see Figure~\ref{fig:knowledge_method}: a (top)) remain limiting factors. First, random initialization may cause prompt tokens to fall into semantically irrelevant regions, leading to an optimization starting point misaligned with the target semantic space and increasing the risk of converging to suboptimal solutions. Second, as training progresses, the static representation of prior knowledge and dynamically optimized learnable tokens tend to exhibit representation drift. This shift becomes particularly pronounced in few-shot scenarios, where the model excessively relies on learnable tokens to fit base class features, failing to effectively leverage prior knowledge for representation adjustment, ultimately degrading cross-class generalization. Furthermore, although learnable tokens and prior knowledge may interact implicitly in subsequent network layers (e.g., Transformer self-attention), the lack of explicit constraints makes it difficult to suppress semantic noise and enhance semantic relevance in the high-dimensional embedding space. Consequently, these limitations still hinder the model’s ability to extract discriminative and generalized features, constraining its performance on both base and novel classes.

To address the aforementioned issues, we propose InPK (\textbf{In}fusing \textbf{P}rior \textbf{K}nowledge into Prompt), which enhances the generalization capability of vision-language models through a hierarchical prior knowledge infusion mechanism. During initialization, we inject class-specific prior knowledge generated by a large language model (LLM) into the learnable tokens, establishing class-related semantic preferences and constraining the tokens within a semantically meaningful subspace (see Figure \ref{fig:knowledge_method}: a (bottom)). For base classes, this design encourages the model to focus on essential features aligned with prior knowledge, enhancing sensitivity to subtle distinctions. For novel classes, the learnable tokens create cross-class semantic bridges through shared visual concepts, allowing the feature representations of novel classes to adaptively integrate into the existing semantic space, thereby improving generalization to unseen classes~\cite{zhang2024concept}. Visualization shows that compared with the method of directly concatenating prior knowledge and learnable tokens, the feature space produced by InPK has tighter intra-class clustering and greater inter-class separation (see Figure \ref{fig:knowledge_method}: b), verifying the effectiveness of our prior knowledge fusion strategy (see Figure \ref{fig:knowledge_method}: c).

Furthermore, to counteract the gradual dilution of class information across multi-layer encoders, we continuously reinforce the interaction between the learnable tokens and prior knowledge at various feature levels, progressively capturing fine-grained vision-semantic associations. For more precise semantic alignment between textual and visual modalities, we introduce a learnable cross-modal projection layer. Moreover, InPK improves model robustness via a regularization mechanism by incorporating a Class-Name Saliency Regularization term into the loss function, encouraging the model to focus on core class semantics while mitigating over-reliance on shared attributes. Notably, experimental results of average accuracy on 11 popular recognition datasets show that InPK significantly outperforms state-of-the-art methods on the base and novel classes. 

The main contributions of this paper include:
\begin{itemize}
     \item 
     We infuse prior knowledge into learnable tokens at multiple feature levels, guiding the model to extract more discriminative and generalizable features, thereby enhancing recognition performance on both base and novel classes.
     \item We introduce a learnable text-to-vision projection layer to accommodate the text prompt tuning process, thereby enhancing vision-text semantic alignment and improving cross-modal consistency.
    \item We conduct extensive experiments on 11 popular vision recognition datasets, and the results demonstrate that InPK achieves significant performance improvements across multiple zero/few-shot image classification tasks.
\end{itemize}

\section{Related Work}
\label{sec:formatting}

\subsection{Vision-Language Models}
Vision-Language Models (VLMs) represent significant progress in multimodal learning~\cite{wei2021meta,9454290,wei2020universal} in recent years. These models are pre-trained on large-scale datasets to effectively learn joint representations from both images and text. Recent studies~\cite{radford2021learning,jia2021scaling,yao2021filip,yuan2021florence,liu2023learning} have demonstrated VLMs' superior performance in tasks such as zero/few-shot image recognition. Notably, the pioneering work CLIP ~\cite{radford2021learning} is renowned for its simplicity and effectiveness, leveraging large-scale image-text pair training and contrastive learning techniques. Other works like ALIGN~\cite{jia2021scaling}, FILIP~\cite{yao2021filip}, Florence~\cite{yuan2021florence}, and REACT~\cite{liu2023learning} further highlight VLMs' strong open vocabulary understanding. To adapt pre-trained VLMs to specific downstream tasks, numerous task-specific methods have been proposed including segmentation~\cite{wysoczanska2024clip,zhang2024exploring,zhou2022extract,wang2024cm}, image recognition~\cite{zhang2021tip,alayrac2022flamingo,yao2023visual,yang2024mma,wei2024runge, wu2024fine} and object detection~\cite{fang2024simple,zang2022open,zhou2022detecting,zhong2022regionclip,ke2024vldadaptor}. In this work, we propose a novel prompt-tuning technique that enhances VLMs’ generalization performance across various visual recognition tasks.

\subsection{Prompt Learning for Vision-Language Models}
Fine-tuning Vision-Language Models (VLMs) for downstream tasks or datasets while preserving their original generalization capabilities poses significant challenges. Retraining the entire model is often impractical due to the vast number of parameters, and it risks overfitting, which could diminish the generalization benefits obtained during pre-training. To tackle this, prompt-tuning methods have emerged, with CoOp ~\cite{zhou2022learning} being a pioneering approach. CoOp adapts VLMs to downstream tasks by introducing learnable vectors in place of hand-crafted prompts. Extensions of this approach, multimodal prompt tuning~\cite{khattak2023maple,liu2024multi,cho2023distribution,zhang2024unleash} has been applied to fine-tune both text and visual branches simultaneously. To prevent overfitting, a line of works\cite {khattak2023self,yao2023visual,yao2024tcp,zhu2023prompt} have introduced regularization constraints to reduce the loss of general information. Other approaches, such as UNIGRAM~\cite{li2023gradient} and ProMetaR~\cite{park2024prompt}, leverage meta-learning to enhance generalization by initializing prompts or applying regularization in the meta-learning framework. More recently, some methods~\cite{lee2024coapt,menon2022visual,wang2024learning,zhang2024concept,kan2023knowledge} have introduced external knowledge to enrich textual representations and better capture complex visual semantics. Our approach goes beyond using learnable tokens by infusing them into class-specific prior knowledge to provide class-relevant preferences, resulting in improved generalization performance on downstream tasks.

\section{Methodology}
In this section, we first provide a brief overview of CLIP~\cite{radford2021learning}, followed by a comprehensive introduction to our proposed method, InPK, including the generation of prior knowledge, the prior knowledge-infused text encoder, the alignment of text and visual branches, and the training objective.

\subsection{Review of CLIP}
We build our method on a pre-trained vision-language model CLIP. After feeding the CLIP model an image and its corresponding text description, CLIP employs a contrastive loss to establish global alignment between image and text. CLIP comprises a text encoder \(\theta\) and an image encoder \(\phi\), which work together to map the image and text into a shared feature space.

In the visual branch, the input image $\mathcal{I}$ is divided into patches, resulting in \( \boldsymbol{U}=\{ \boldsymbol{v_{\text{cls}}, v} \} \), where \( \boldsymbol{v_{\text{cls}}} \) is the class token and \( \boldsymbol{v} \) represents the patch embeddings. Subsequently, the sequence \(\boldsymbol{U} \) is passed through the image encoder to obtain the visual feature \( \boldsymbol{x}=\phi \boldsymbol{(U)}\), where $\boldsymbol{x} \in \mathbb{R}^{d}$. 

In the text branch, the input is defined as \(\boldsymbol{T}= \{ \boldsymbol{T_0}, \boldsymbol{c_k} \}\), where \(\boldsymbol{T_0}\) represents hand-crafted templates like ``a photo of a \(\{\}\)", and \( \boldsymbol{c_k} \) denotes the class name for the \(k\)-th class. The input \(\boldsymbol{T}\) is tokenized into \(n\) word embeddings \(\boldsymbol{E}= \{ \boldsymbol{e_1}, \boldsymbol{e_2}, \ldots, \boldsymbol{e_n} \}\), which are then fed sequentially into \(L\) transformer layers \(\{\theta_i\}_{i=1}^{L}\):

\begin{equation}
\begin{aligned}
\boldsymbol{{E}^{i}} = \theta_{i}(\boldsymbol{E^{i-1}}) \quad  i=1, 2, \cdots, L.
\end{aligned}
\end{equation}
Then, the text embedding \(\boldsymbol{e_n^{L}}\) then corresponding to the final token of the last transformer layer is projected into the common vision-language space to obtain the feature \(\boldsymbol{f} \in \mathbb{R}^{d}\):

\begin{equation}
\begin{aligned}
\boldsymbol{f} & =\textit{Proj}\left(\boldsymbol{{{e}}_n^{L}}\right).
\end{aligned}
\end{equation}

Finally, for zero-shot inference, text prompts are provided with class labels \( y \in \{1,2, \ldots, N_c\} \), and the text feature \( \boldsymbol{f} \)  is matched with the image  feature \( \boldsymbol{x} \) using the following: 
\begin{equation}
p( \hat{y}=i \mid \mathcal{I})=\frac{\exp \left(\operatorname{sim}\left(\boldsymbol{x}, \boldsymbol{f_i}\right) / \tau\right)}{\sum_{k=1}^{N_c} \exp \left(\operatorname{sim}\left(\boldsymbol{x}, \boldsymbol{f_k}\right) / \tau\right)},
\end{equation}
where $\operatorname{sim}(\cdot, \cdot)$ denotes cosine similarity, and \(\tau\) is the temperature.

\subsection{InPK: Infusing Prior Knowledge into Prompt}


\begin{figure*}[h]
    \centering
    \includegraphics[width=1\linewidth]{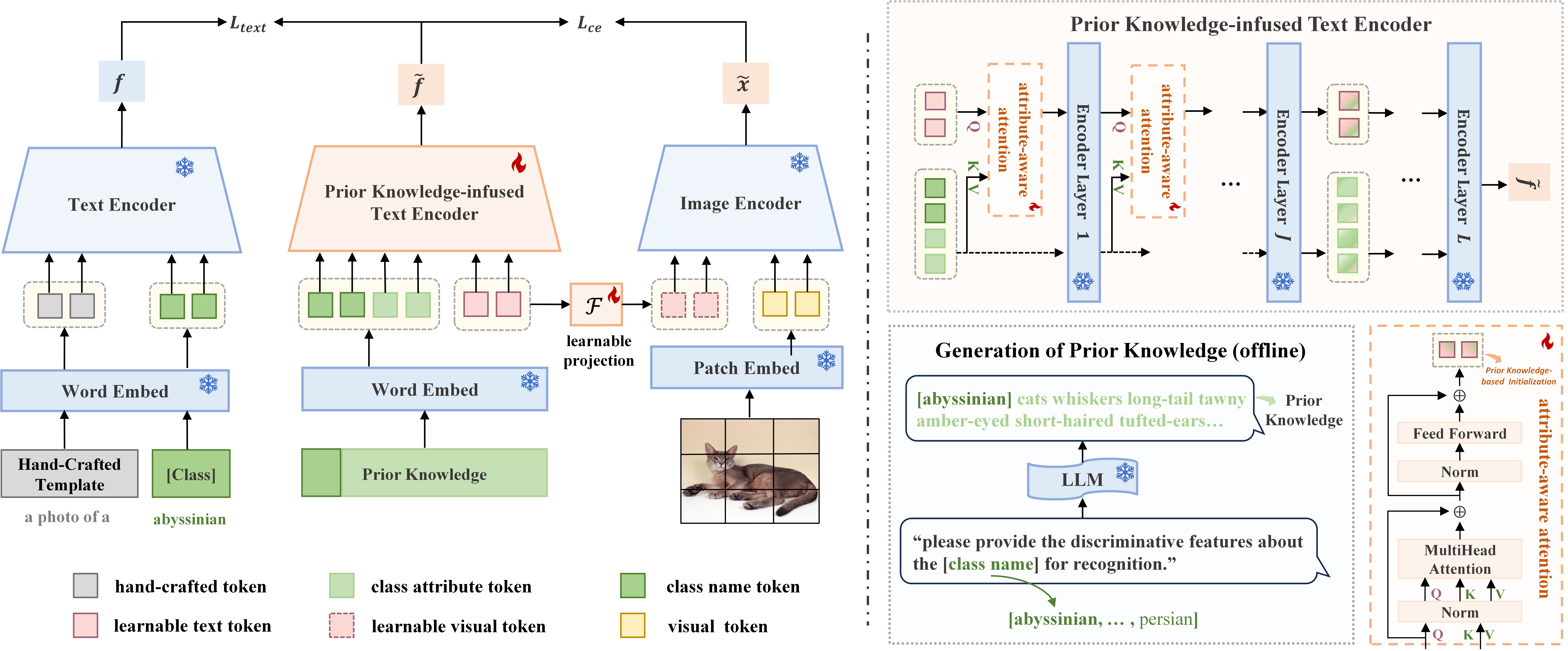}
     \vspace{-10pt}
    \caption{Overview of InPK method. Prior knowledge is generated offline using a predefined instruction template and subsequently fed into the \textbf{P}rior \textbf{K}nowledge-\textbf{i}nfused text encoder (PKi). Within PKi, class-specific prior knowledge is infused into learnable tokens through attribute-aware attention at the initialization stage, and the interaction between tokens and prior knowledge is progressively reinforced across multiple feature levels. Meanwhile, we introduce a learnable text-to-vision projection layer to better align visual-text semantics. Furthermore,  loss \( L_{text}\) is applied to mitigate the model's forgetting of general information and to emphasize the role of class names.
}
    \vspace{-10pt}
    \label{fig:overview}
\end{figure*}
Based on the CLIP model, to further adapt the VLMs to specific downstream tasks while excelling in both base and novel classes, we propose the InPK model, as depicted in Figure ~\ref{fig:overview}. Different from other methods based on prior knowledge~\cite{lee2024coapt,wang2024learning,royconsistency}, which randomly initialize the learnable tokens, our approach infuses prior knowledge into learnable tokens at the initialization stage and progressively strengthens the connection between prior knowledge and tokens across multiple feature levels. Furthermore, we introduce a learnable text-to-vision projection layer to accommodate text adjustments and enforce alignment between visual and textual semantics. Finally, regularization constraint is employed with the objective of retaining general knowledge and emphasizing the role of class names.

\subsubsection{Generation of Prior Knowledge}
Regarding prior knowledge, relying solely on class names, especially for rare or ambiguous scientific terms, limits the model's ability to effectively associate these terms with visual information. Therefore, it is essential to incorporate attribute words related to each category alongside the class names~\cite{kan2023knowledge}. In our method, prior knowledge includes both class names and class-specific attributes, enriching the model's semantic understanding and universal visual concepts brought by attributes that provide cross-class knowledge. We generate attribute words using GPT-4, leveraging its rich world knowledge, and guide the process with a predefined instruction template that incorporates both the class name and dataset name. The template is as follows: “Please provide the discriminative features about the [class name] for recognition”. This process yields the input \(\boldsymbol{G} = \{\boldsymbol{c_k, g_1, \dots, g_N}\}\), where \( \boldsymbol{c_k} \) denotes the class name for the \(k\)-th class, and \(\{\boldsymbol{g_1, \dots, g_N}\}\) represent \(N\) attribute words. Prior knowledge is generated offline, eliminating the need for training or inference time. 

\subsubsection{Prior Knowledge-infused Text Encoder}
With the acquired class-specific prior knowledge, we avoid merely concatenating the learnable tokens with prior knowledge before feeding them into the text encoder. Instead, our approach infuses prior knowledge into the learnable tokens, introducing explicit category preference and delivering higher-quality input to subsequent layers. Therefore, beyond the original feature \( \boldsymbol{f} \) derived from hand-crafted prompts in the text branch, we introduce an additional pathway within the Prior Knowledge-infused text encoder(PKi) that incrementally models features, ultimately producing the new textual feature \( \boldsymbol{ \widetilde{f}}\). Specifically, the input tokens for the PKi module are structured as \( \{ \boldsymbol{P_t,\widetilde{E}} \} \), where \( \boldsymbol{P_t} = \{ \boldsymbol{p_t^1, p_t^2, \dots, p_t^M} \} \) consists of \( M \) learnable tokens, and \(\boldsymbol{ \widetilde{E}} = \{ \boldsymbol{ \widetilde{e}_1,  \widetilde{e}_2, \ldots,  \widetilde{e}_n} \}\) represents the word embeddings of \(\boldsymbol{G}\). We then introduce an attribute-aware attention mechanism \( \mathcal{T} \) that fuses class-specific knowledge with the learnable tokens. In this mechanism, the learnable tokens \( \boldsymbol{P_t} \) serve as the query \( Q \), while the class-specific prior knowledge embeddings \( \boldsymbol{\widetilde{E}} \) as both the key \( K \) and the value \( V \), resulting in a robust token embedding \( \boldsymbol{ \widehat{P_t}} \) that incorporates prior knowledge:
\begin{equation}
\begin{aligned}
\boldsymbol{\widehat{P_t}}=\mathcal{T}(\boldsymbol{P_t}, \boldsymbol{\widetilde{E}})=\textit{FFN}(\textit{LN}(\boldsymbol{A}))+\boldsymbol{A},
\end{aligned}
\end{equation}
\begin{equation}
\begin{aligned}
\boldsymbol{A}=\textit{MultiHead}(\textit{LN}(\boldsymbol{P_t}), \textit{LN}(\boldsymbol{\widetilde{E}}), \textit{LN}(\boldsymbol{\widetilde{E}}))+\boldsymbol{P_t},
\end{aligned}
\end{equation}
where $\textit{MultiHead ()}$ and $\textit{FFN ()}$ follow the standard Transformer configurations~\cite{vaswani2017attention}, representing multi-head attention and feed-forward neural network, respectively. Layer normalization ($\textit{LN}$) is applied before each block. 

Furthermore, to counteract the potential dilution of category information caused by multi-layer encoders, we incorporate attribute-aware attention into each transformer layer of the PKi up to a specific depth \( J \), allowing for step-by-step feature modeling. Specifically, the PKi itself comprises \( L \) layers of frozen transformers, with all parameters pre-trained. The fused learnable token \( \boldsymbol{ \widehat{P}_{t}^{i-1}} \) is concatenated with the \( \boldsymbol{\widetilde{E}^{i-1}} \), and the resulting tokens are then passed to the $i^{\mathrm{th}}$ transformer layer:
\begin{equation}
\begin{aligned}
\boldsymbol{\widehat{P}_t^{i-1}}=\mathcal{T}(\boldsymbol{P_t^{i-1}}, \boldsymbol{\widetilde{E}^{i-1}}),
\end{aligned}
\end{equation}
\begin{equation}
\begin{aligned}
[\boldsymbol{{P_t^i},  \widetilde{E}^{i}}] = \theta_{i}([\boldsymbol{{ \widehat{P}}_t^{i-1},   \widetilde{E}^{i-1}}]) \quad  i=1, 2, \cdots, J,
\end{aligned}
\end{equation}
where $\operatorname[\cdot, \cdot]$ refers to concatenation operation and the $\theta_{i}$ is the $i^{\mathrm{th}}$ transformer layer in text Encoder. After $J^{\mathrm{th}}$  transformer layer, the tokens are fed into the remaining layers and obtain the final text feature \( \boldsymbol{ \widetilde{f}}\in \mathbb{R}^{d}\):
\begin{equation}
\begin{aligned}
[\boldsymbol{{P_t^j},  \widetilde{E}^{j}}] = \theta_{j}([\boldsymbol{P_t^{j-1}},  \boldsymbol{ \widetilde{E}^{j-1}}]) \quad j=J+1, \cdots, L,
\end{aligned}
\end{equation}
\begin{equation}
\begin{aligned}
\boldsymbol{ \widetilde{f}} & =\textit{Proj}\left(\boldsymbol{{ \widetilde{e}}_n^{L}}\right).
\end{aligned}
\end{equation}

\subsubsection{Alignment of Text and Visual Branches}
Due to the adjustment of the text branch, where additional prior knowledge is introduced, we propose a trainable text-to-vision projection layer 
\(\mathcal{F}\) to better align visual and textual semantics. The textual learnable tokens \( \boldsymbol{P_t} \) are passed through this projection layer to generate the corresponding visual learnable tokens \( \boldsymbol{P_v} \). The text and visual branches are dynamically adjusted using cross-modal shared parameters, ensuring consistent alignment between the two modalities. The specific formula is as follows:
\begin{equation}
\boldsymbol{P_v} =\mathcal{F}(\boldsymbol{P_t}),
\end{equation}
where \( \boldsymbol{P_v} = \{ \boldsymbol{p_v^1}, \boldsymbol{p_v^2}, \dots, \boldsymbol{p_v^M} \} \) consists of \( M \) learnable visual tokens. By combining the fixed token \( \boldsymbol{U} \) obtained from CLIP, we define the token of the new visual branch as \( \boldsymbol{V} = \{ \boldsymbol{U}, \boldsymbol{P_v} \} \). Then, the visual token \( \boldsymbol{V} \) is passed through the image encoder to obtain new visual features \( \boldsymbol{ \widetilde{x}} = \phi (\boldsymbol{V}) \), where \( \boldsymbol{ \widetilde{x}}\in \mathbb{R}^{d}\). 

\subsubsection{Training Objective}
In the new text and visual branches, text features \( \boldsymbol{ \widetilde{f}}\) and visual features \( \boldsymbol{ \widetilde{x}}\) are obtained. Thus, the prediction probability for $\mathcal{I}$ pertaining to label \( i\) is represented as:
\begin{equation}
p( \widehat{y}=i \mid \mathcal{I})=\frac{\exp \left(\operatorname{sim}\left(\boldsymbol{ \widetilde{x}}, \boldsymbol{ \widetilde{f}_i}\right) / \tau\right)}{\sum_{k=1}^{N_c} \exp \left(\operatorname{sim}\left(\boldsymbol{ \widetilde{x}}, \boldsymbol{ \widetilde{f}_k}\right) / \tau\right)},
\end{equation}
where $\operatorname{sim}(\cdot, \cdot)$ denotes the cosine similarity, \( \tau \) is a temperature parameter that controls the sharpness of the softmax distribution, and \( N_c \) is the number of seen classes. Thus, the cross-entropy loss between the true label and the predicted label is minimized as follows:
\begin{equation}
\mathcal{L}_{ce}=-\frac{1}{N} \sum^N_{i=1} \log \left(P\left( \hat{y}=y_i \mid \mathcal{I}_i\right)\right),
\end{equation}
where \( N \) represents the number of samples in the training dataset, \( y_i \) denotes the ground-truth label of image \( \mathcal{I}_i \).

Furthermore, the attribute words of different categories are often generic. For example, Felidae species such as the cat and tiger share common fine-grained attributes, including the presence of whiskers, a long tail, and so on. The constraints of $\mathcal{L}_{ce}$ alone would cause the prompt to become largely dependent on the attribute words, potentially confusing the categories. It is therefore necessary to enhance the role of class names by leveraging the differences of class names within the general text features  \( \boldsymbol{f} \). Following the PromptSRC~\cite{khattak2023self} approach, we introduce a Class-Name Saliency Regularization constraint using the \( L1 \) Loss on the prompted textual features \( \boldsymbol{\widetilde{f}} \) to ensure their consistency with the CLIP pre-trained features \( \boldsymbol{f} \). This regularization prevents the model from overfitting to overly generic or misleading attributes, ensuring that the generated features align with the pre-trained model’s feature space without consuming additional inference time. The loss is minimized as follows:
\begin{equation}
\mathcal{L}_{text}=\sum_{i=1}^d\left| \widetilde{\boldsymbol{f}}-{\boldsymbol{f}}\right|.
\end{equation}



The overall training objective becomes:
\begin{equation}
\mathcal{L}=\mathcal{L}_{ce}+\lambda \mathcal{L}_{text}.
\end{equation}

\section{Experiments}\label{par: 6}
\subsection{Experimental Setup}
We follow the experimental setting~\cite{zhou2022learning} of previous methods to conduct experiments through four evaluation protocols: base-to-novel generalization, few-shot classification, cross-dataset evaluation, and domain generalization.

\begin{table*}[t]
  \caption{Comparison with CoOp, CoCoOp and TCP in the base-to-novel generalization setting. HM: harmonic mean. The best results for each column are highlighted in \textbf{bold} font, and the improvement of our method over the SOTA method TCP is indicated as $\Delta$. Our method exhibits outstanding performance in both base and novel classes.}
 \label{tab:base-to-new}
  \begin{minipage}[t]{0.31\linewidth}
    \centering
    
    \caption*{   \centering  (a) \textbf{Average results }} 
    \scalebox{1.2}{

  \begin{tabular}{@{\hspace{7pt}}c@{\hspace{7pt}}c@{\hspace{7pt}}c@{\hspace{7pt}}c}
       \hline\noalign{\smallskip}
        Method & Base &  {Novel} & HM \\
    \hline\noalign{\smallskip}
            CoOp    & 82.69 & 63.22 & 71.66 \\
            CoCoOp  & 80.47 & 71.69 & 75.83 \\
            TCP & 84.13 & 75.36 & 79.51 \\
         \hline\noalign{\smallskip}
     \cellcolor{lightgray!30}Ours & \cellcolor{lightgray!30}\textbf{85.17} & \cellcolor{lightgray!30}\textbf{78.79} & \cellcolor{lightgray!30}\textbf{81.85} \\
            $\Delta$ &\textcolor[rgb]{ .184,  .459,  .71}{+1.04}  &\textcolor[rgb]{ .184,  .459,  .71}{+3.43} & \textcolor[rgb]{ .184,  .459,  .71}{+2.34} \\
            \hline
    \end{tabular}
    }
    \vspace{+0.4cm}

      \caption*{  \centering  (d) DTD} 
\scalebox{1.2}{
  \begin{tabular}{@{\hspace{7pt}}c@{\hspace{7pt}}c@{\hspace{7pt}}c@{\hspace{7pt}}c}
            \hline\noalign{\smallskip}
            Method & Base  & Novel & HM \\
            \hline\noalign{\smallskip}
            CoOp & 79.44 & 41.18 & 54.24 \\ 
            CoCoOp & 77.01 & 56.00 & 64.85 \\
           TCP & 82.77 & 58.07 & 68.25 \\
            \hline\noalign{\smallskip}
            \cellcolor{lightgray!30}Ours & \cellcolor{lightgray!30}\textbf{84.43} & \cellcolor{lightgray!30}\textbf{71.30} & \cellcolor{lightgray!30}\textbf{77.31} \\
            $\Delta$ &\textcolor[rgb]{ .184,  .459,  .71}{+1.66}  &\textcolor[rgb]{ .184,  .459,  .71}{+13.23} & \textcolor[rgb]{ .184,  .459,  .71}{+9.06} \\
            \hline
      \end{tabular}
      }
\vspace{+0.4cm}

    \caption*{    \centering (g) OxfordPets} 
  \scalebox{1.2}{

      \begin{tabular}{@{\hspace{7pt}}c@{\hspace{7pt}}c@{\hspace{7pt}}c@{\hspace{7pt}}c}
            \hline\noalign{\smallskip}
            Method & Base  & Novel & HM \\
            \hline\noalign{\smallskip}
            CoOp & 93.67 & 95.29 & 94.47 \\
            CoCoOp & \textbf{95.20} & \textbf{97.69} & \textbf{96.43} \\
            TCP & 94.67 & 97.20 & 95.92 \\
            \hline\noalign{\smallskip}
            \cellcolor{lightgray!30}Ours & \cellcolor{lightgray!30}94.67 & \cellcolor{lightgray!30}97.63 & \cellcolor{lightgray!30}96.13 \\
            $\Delta$ &\textcolor[rgb]{ .184,  .459,  .71}{+0.00}  &\textcolor[rgb]{ .184,  .459,  .71}{+0.43} & \textcolor[rgb]{ .184,  .459,  .71}{+0.21} \\
            \hline
    \end{tabular}
    }
     \vspace{+0.4cm}
    \caption*{   \centering (j) Food101} 
 \scalebox{1.2}{

  \begin{tabular}{@{\hspace{7pt}}c@{\hspace{7pt}}c@{\hspace{7pt}}c@{\hspace{7pt}}c}
            \hline\noalign{\smallskip}
            Method & Base  & Novel & HM \\
            \hline\noalign{\smallskip}
            CoOp & 88.33 & 82.26 & 85.19 \\
            CoCoOp & \textbf{90.70} & 91.29 & \textbf{90.99} \\
            TCP & 90.57& 91.37 & 90.97  \\
            \hline\noalign{\smallskip}
            \cellcolor{lightgray!30}Ours & \cellcolor{lightgray!30}90.10 & \cellcolor{lightgray!30}\textbf{91.57} & \cellcolor{lightgray!30}90.83 \\
            $\Delta$ &\textcolor[rgb]{ .776,  .349,  .067}{-0.47}  &\textcolor[rgb]{ .184,  .459,  .71}{+0.20} & \textcolor[rgb]{ .776,  .349,  .067}{-0.14} \\
            \hline
    \end{tabular}
    }
     \vspace{+0.4cm}
   
  \end{minipage}
  \hspace{0.5cm}
  \begin{minipage}[t]{0.31\linewidth}
    \centering
   
     \caption*{   \centering (b) ImageNet} 
\scalebox{1.2}{

 \begin{tabular}{@{\hspace{7pt}}c@{\hspace{7pt}}c@{\hspace{7pt}}c@{\hspace{7pt}}c}
            \hline\noalign{\smallskip}
            Method & Base  & Novel & HM \\
            \hline\noalign{\smallskip}
            CoOp & 76.47 & 67.88 & 71.92 \\
            CoCoOp & 75.98 & 70.43 & 73.10 \\
            TCP & 77.27 & 69.87 & 73.38 \\
            \hline\noalign{\smallskip}
            \cellcolor{lightgray!30}Ours & \cellcolor{lightgray!30}\textbf{77.60} & \cellcolor{lightgray!30}\textbf{71.37} & \cellcolor{lightgray!30}\textbf{74.35} \\
            $\Delta$ &\textcolor[rgb]{ .184,  .459,  .71}{+0.33}  &\textcolor[rgb]{ .184,  .459,  .71}{+1.50} & \textcolor[rgb]{ .184,  .459,  .71}{+0.97}  \\
            \hline
    \end{tabular}
    }
    \vspace{+0.4cm}

   \caption*{   \centering (e) EuroSAT} 
\scalebox{1.2}{
 \begin{tabular}{@{\hspace{7pt}}c@{\hspace{7pt}}c@{\hspace{7pt}}c@{\hspace{7pt}}c}
            \hline\noalign{\smallskip}
            Method & Base  & Novel & HM \\
            \hline\noalign{\smallskip}
            CoOp & 92.19 & 54.74 & 68.69 \\
            CoCoOp & 87.49 & 60.04 & 71.21 \\
            TCP & 91.63 & 74.73 & 82.32 \\
            \hline\noalign{\smallskip}
            \cellcolor{lightgray!30}Ours & \cellcolor{lightgray!30}\textbf{95.70} & \cellcolor{lightgray!30}\textbf{82.37} & \cellcolor{lightgray!30}\textbf{88.54}\\
            $\Delta$ &\textcolor[rgb]{ .184,  .459,  .71}{+4.07}  &\textcolor[rgb]{ .184,  .459,  .71}{+7.64} & \textcolor[rgb]{ .184,  .459,  .71}{+6.21} \\
            \hline
      \end{tabular}

      }
\vspace{+0.4cm}

    \caption*{   \centering (h) StanfordCars} 
  \scalebox{1.2}{
  
  \begin{tabular}{@{\hspace{7pt}}c@{\hspace{7pt}}c@{\hspace{7pt}}c@{\hspace{7pt}}c}
            \hline\noalign{\smallskip}
            Method & Base  & Novel & HM \\
            \hline\noalign{\smallskip}
            CoOp & 78.12 & 60.40 & 68.13 \\
            CoCoOp & 70.49 & 73.59 & 72.01 \\
            TCP & 80.80 & 74.13 & 77.32 \\
            \hline\noalign{\smallskip}
            \cellcolor{lightgray!30}Ours & \cellcolor{lightgray!30}\textbf{81.97} & \cellcolor{lightgray!30}\textbf{76.10} & \cellcolor{lightgray!30}\textbf{78.93} \\
           $\Delta$ &\textcolor[rgb]{ .184,  .459,  .71}{+1.17}  &\textcolor[rgb]{ .184,  .459,  .71}{+1.97} & \textcolor[rgb]{ .184,  .459,  .71}{+1.60} \\
            \hline
    \end{tabular}
    }
     \vspace{+0.4cm}
    \caption*{   \centering (k) FGVCAircraft} 
  \scalebox{1.2}{
  
  \begin{tabular}{@{\hspace{7pt}}c@{\hspace{7pt}}c@{\hspace{7pt}}c@{\hspace{7pt}}c}
            \hline\noalign{\smallskip}
            Method & Base  & Novel & HM \\
            \hline\noalign{\smallskip}
            CoOp & 40.44 & 22.30 & 28.75 \\
            CoCoOp & 33.41 & 23.71 & 27.74 \\
            TCP & 41.97 & 34.43 & 37.83 \\
            \hline\noalign{\smallskip}
            \cellcolor{lightgray!30}Ours & \cellcolor{lightgray!30}\textbf{45.67} & \cellcolor{lightgray!30}\textbf{36.50} & \cellcolor{lightgray!30}\textbf{40.57}  \\
            $\Delta$ &\textcolor[rgb]{ .184,  .459,  .71}{+3.70}  &\textcolor[rgb]{ .184,  .459,  .71}{+2.07} & \textcolor[rgb]{ .184,  .459,  .71}{+2.75} \\
            \hline
    \end{tabular}
    }
     \vspace{+0.4cm}

  \end{minipage}
  \hspace{0.5cm}
  \begin{minipage}[t]{0.31\linewidth}
    \centering

    \caption*{   \centering    (c) Caltech101} 
 \scalebox{1.2}{
   \begin{tabular}{@{\hspace{7pt}}c@{\hspace{7pt}}c@{\hspace{7pt}}c@{\hspace{7pt}}c}
            \hline\noalign{\smallskip}
            Method & Base  & Novel & HM \\
            \hline\noalign{\smallskip}
            CoOp & 98.00 & 89.81 & 93.73 \\
            CoCoOp & 97.96 & 93.81 & 95.84 \\
           TCP & 98.23 & \textbf{94.67} & 96.42 \\
            \hline\noalign{\smallskip}
            \cellcolor{lightgray!30} Ours & \cellcolor{lightgray!30}\textbf{98.67} & \cellcolor{lightgray!30}94.30 & \cellcolor{lightgray!30}\textbf{96.44} \\
            $\Delta$ &\textcolor[rgb]{ .184,  .459,  .71}{+0.44}  &\textcolor[rgb]{ .776,  .349,  .067}{-0.37} & \textcolor[rgb]{ .184,  .459,  .71}{+0.02}  \\
            \hline
    \end{tabular}
    }
     \vspace{+0.4cm}

\caption*{   \centering (f) UCF101} 
  \scalebox{1.2}{
  
  \begin{tabular}{@{\hspace{7pt}}c@{\hspace{7pt}}c@{\hspace{7pt}}c@{\hspace{7pt}}c}
            \hline\noalign{\smallskip}
            Method & Base  & Novel & HM \\
            \hline\noalign{\smallskip}
            CoOp & 84.69 & 56.05 & 67.46 \\
            CoCoOp & 82.33 & 73.45 & 77.67 \\
             TCP & 87.13 & 80.77 & 83.83 \\
            \hline\noalign{\smallskip}
            \cellcolor{lightgray!30}Ours & \cellcolor{lightgray!30}\textbf{87.13} & \cellcolor{lightgray!30}\textbf{83.10} & \cellcolor{lightgray!30}\textbf{85.07} \\
            $\Delta$ &\textcolor[rgb]{ .184,  .459,  .71}{+0.00}  &\textcolor[rgb]{ .184,  .459,  .71}{+2.33} & \textcolor[rgb]{ .184,  .459,  .71}{+1.24} \\
            \hline
    \end{tabular}
    }

    \vspace{+0.4cm}

    \caption*{   \centering (i) Flowers102} 
\scalebox{1.2}{

  \begin{tabular}{@{\hspace{7pt}}c@{\hspace{7pt}}c@{\hspace{7pt}}c@{\hspace{7pt}}c}
            \hline\noalign{\smallskip}
            Method & Base  & Novel & HM \\
            \hline\noalign{\smallskip}
            CoOp & 97.60 & 59.67 & 74.06 \\
            CoCoOp & 94.87 & 71.75 & 81.71 \\
            TCP & 97.73 & 75.57 & 85.23 \\
            \hline\noalign{\smallskip}
            \cellcolor{lightgray!30}Ours & \cellcolor{lightgray!30}\textbf{98.17} & \cellcolor{lightgray!30}\textbf{81.27} & \cellcolor{lightgray!30}\textbf{88.92} \\
           $\Delta$ &\textcolor[rgb]{ .184,  .459,  .71}{+0.44}  &\textcolor[rgb]{ .184,  .459,  .71}{+5.70} & \textcolor[rgb]{ .184,  .459,  .71}{+3.69} \\
            \hline
    \end{tabular}
    }
 \vspace{+0.4cm}
    \caption*{   \centering (l) SUN397} 
  \scalebox{1.2}{
  
  \begin{tabular}{@{\hspace{7pt}}c@{\hspace{7pt}}c@{\hspace{7pt}}c@{\hspace{7pt}}c}
            \hline\noalign{\smallskip}
            Method & Base  & Novel & HM \\
            \hline\noalign{\smallskip}
            CoOp & 80.60 & 65.89 & 72.51 \\
            CoCoOp & 79.74 & 76.86 & 78.27 \\
            TCP & 82.63 & 78.20 & 80.35 \\
            \hline\noalign{\smallskip}
            \cellcolor{lightgray!30}Ours & \cellcolor{lightgray!30}\textbf{82.77} & \cellcolor{lightgray!30}\textbf{81.13} & \cellcolor{lightgray!30}\textbf{81.94} \\
            $\Delta$ &\textcolor[rgb]{ .184,  .459,  .71}{+0.14}  &\textcolor[rgb]{ .184,  .459,  .71}{+2.93} & \textcolor[rgb]{ .184,  .459,  .71}{+1.59} \\
            \hline
    \end{tabular}
    }
\vspace{+0.4cm}
 \end{minipage}
       \vspace{-8pt}
\end{table*}

\begin{table*}
\caption{ Comparison between InPK and the existing state-of-the-art methods in the base-to-novel generalization setting. Our method achieves the best accuracy in both base and novel classes, demonstrating the strong generalization ability of our method.} 
\centering
\renewcommand{\arraystretch}{1.2}
 \setlength{\tabcolsep}{4pt}

    \scalebox{1.1}[1.1]{
    \begin{tabular}{lcccccccccccc}
    \toprule
       & CoOp~\cite{zhou2022learning} & CoCoOp~\cite{zhou2022conditional} & PromptSRC~\cite{khattak2023self} & HPT~\cite{wang2024learning} & ProMetaR~\cite{park2024prompt}  & TCP ~\cite{yao2024tcp} & CoPrompt~\cite{royconsistency}  & MMA~\cite{yang2024mma} & \multirow{2}{*}{\textbf{Ours}} \\
     & \small{(IJCV22)} & \small{(CVPR22)} & \small{(ICCV23)}  & \small{(AAAI24)} & \small{(CVPR24)} & \small{(CVPR24)} & \small{(ICLR24)} & \small{(CVPR24)}  &  \\
    \midrule 
     Base & 82.69 & 80.47 & 84.26  & 84.32 & 84.39 & 84.13  & 84.00 & 83.20 & \textbf{85.17} \\
      Novel  & 63.22 & 71.69 & 76.10 & 76.86 & 76.93 & 75.36   & 77.23 & 76.80 & \textbf{78.79} \\
      HM  & 71.66 & 75.83 & 79.97& 80.23 & 80.49  & 79.51  & 80.48 & 79.87  & \textbf{81.85}\\
      
    \bottomrule
    \end{tabular}
    }  
      
    \label{tab:other_method}
       \vspace{8pt}
\end{table*}

\begin{figure*}[h]
  \centering
       \includegraphics[width=0.95\linewidth]{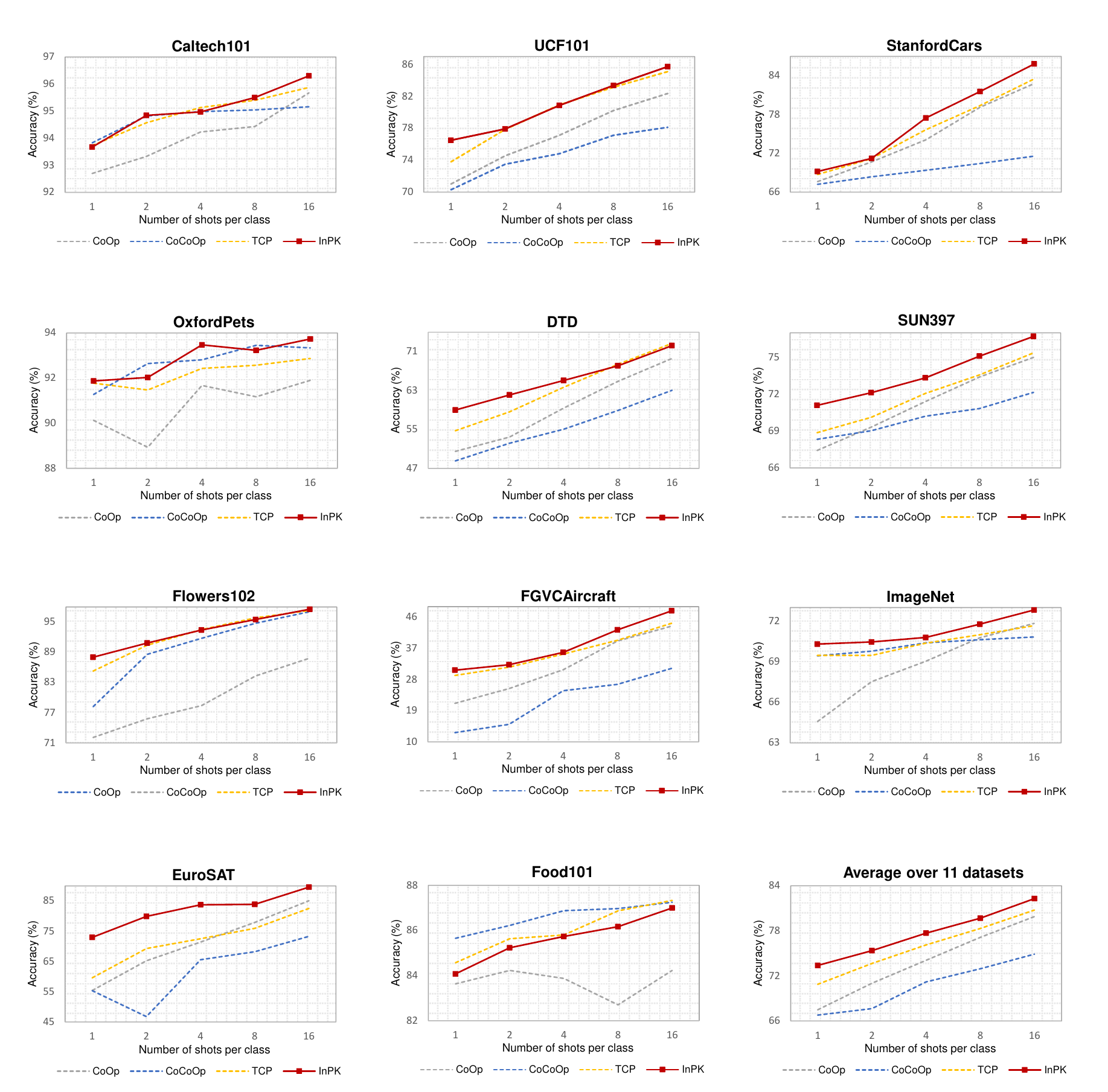}
  \hfill
  
\caption{Comparison of InPK with existing methods in the few-shot learning setting. All models are trained with 1, 2, 4, 8, and 16 shots per class and deployed on the full test set. Our method shows competitive performance in few-shot learning, achieving the highest average results.} 

  \label{fig:few-shot}
       \vspace{-5pt}
\end{figure*}

\begin{table*}[]
 \caption{Comparison of InPK and existing methods in the cross-data evaluation setting. All models are trained with 16-shot samples on the source dataset ImageNet and evaluated on 10 other datasets. Our method achieves the highest average accuracy, indicating that InPK has excellent robustness to different data distributions.}   
    \small \centering
    \renewcommand{\arraystretch}{1.2}
    \scalebox{1}[1]{
    \begin{tabular}{l cccccccccccc}
    \toprule
     \multirow{2.5}{*}{\textbf{Method}}&  \textbf{Source} & \multicolumn{11}{c}{\textbf {Target}} \\ \cmidrule(lr){2-2} \cmidrule(lr){3-13}
    & \textbf{ImNet}  & \textbf {Caltech} &  \textbf{Pets} &  \textbf{Cars} &  \textbf{Flowers} & \textbf {Food} &  \textbf{Aircraft} &  \textbf{SUN397} & \textbf {DTD} & \textbf {EuroSAT} &  \textbf{UCF} & \textbf{\textit{Average}} \\
    \midrule 
    CoOp &   71.51 & 93.70 & 89.14 &64.51 & 68.71 & 85.30 & 18.47 & 64.15 & 41.92 & {{46.39}} & 66.55 & 63.88  \\
    CoCoOp & 71.02 & 94.43 & {90.14} & 65.32 & {71.88} & 86.06 & 22.94 &   {67.36} & 45.73 & 45.37 & 68.21 & 65.74 \\
    PromptSRC & 71.27 &  93.60 & 90.25 & 65.70 & 70.25 & 86.15 & 23.90 & 67.10 &46.87 &45.50 & 68.75  & 65.81\\
    ProMetaR & 71.29& 93.74 & 90.59& 65.83&71.13& 86.39 & 24.78& 67.41& 47.08&45.02 &69.50 & 66.61 \\
    CoPrompt & 70.80 & \textbf{94.50} &   {{90.73}} & 65.67 &  72.30 &   86.43 &   {{24.00}} & 67.57 & {47.07} & 51.90 & {69.73} & 67.00 \\
    MMA & 71.00& 93.80 & 90.30 & \textbf{66.13} &72.07& 86.12 &25.33 &68.17& 46.57&49.24 &68.32 & 66.61 \\
    TCP & 71.40& 93.97 & 91.25& 64.69&71.21& \textbf{86.69} & 23.45& 67.15& 44.35&51.45 &68.73 & 66.29 \\
    \midrule
\textbf{Ours} & \textbf{71.77} &   94.47 & \textbf{91.83} &   {65.20} & \textbf{77.17}&{86.17} & \textbf{{27.07}} & \textbf{{70.60}} & \textbf{59.40} &   \textbf{52.80} & \textbf{74.27} &   \textbf{69.90}  \\  
             \bottomrule

    \end{tabular}
    }
     
    \label{tab:cross-data}
    \vspace{8pt}
\end{table*} 

\noindent \textbf{Base-to-novel generalization.} Following the settings of previous methods, we divide the dataset into base and novel classes. The model is trained on a small number of samples from the base classes, and its performance is evaluated on both the base and novel classes. To assess the trade-off between base and novel class performance and verify the generalization capability of the method, we introduce the harmonic mean (HM)~\cite{xian2017zero}, which balances accuracy across the two class sets.      

\noindent \textbf{Few-shot classification.} To evaluate the model's ability to learn with extremely limited supervision, we assess its performance across few-shot scenarios. Specifically, the model is trained on a small number of labeled images (1, 2, 4, 8, and 16 shots per class) and evaluated on a dataset containing the same classes as those in the training samples. 

\noindent \textbf{Cross-dataset evaluation.} This setting aims to verify the model's zero-shot capability in a cross-dataset context. Specifically, the model undergoes few-shot training on the ImageNet-1K dataset and is then directly evaluated on other datasets without further fine-tuning.

\noindent \textbf{Domain generalization.} To assess the model's robustness on out-of-distribution datasets, we evaluate the model trained on ImageNet-1K across four ImageNet variants that exhibit different types of domain shifts: ImageNet-V2, ImageNet-Sketch, ImageNet-A, and ImageNet-R.

\noindent \textbf{Dataset.} To comprehensively evaluate our model in various settings, we test our method on 11 image classification datasets, covering a range of tasks. These include general object recognition datasets: ImageNet~\cite{deng2009imagenet} and Caltech101~\cite{fei2004learning}; fine-grained classification datasets: OxfordPets~\cite{parkhi2012cats}, StanfordCars~\cite{krause20133d}, Flowers102~\cite{nilsback2008automated}, Food101~\cite{bossard2014food}, and FGVCAircraft~\cite{maji2013fine}; scene recognition dataset: SUN397~\cite{xiao2010sun}; action recognition dataset: UCF101~\cite{soomro2012dataset}; texture classification dataset: DTD~\cite{cimpoi2014describing}; and satellite imagery dataset: EuroSAT~\cite{helber2019eurosat}.

\noindent \textbf{Experiment detail.} Our implementation is based on the code of CoOp~\cite{zhou2022learning}. All experiments are fine-tuned using the CLIP model with a ViT-B/16 backbone where $d=512$, and the results are averaged over three runs with different random seeds (1/2/3). We provide additional implementation details of our method. Our approach is implemented using PyTorch~\cite{paszke2017automatic} and Dassl~\cite{zhou2022domain}, in line with previous prompt-tuning works~\cite{zhou2022learning,yao2023visual,zhou2022conditional}. Prior knowledge is generated using GPT-4, with the number of attribute words $N$ set to 25. The text prompt length $M$ is fixed at 6 and initialized as ``X X X X X X \(\{\}\)". The attribute-aware attention module is inserted into the encoder's first  $J=9$ transformer layers. Following the method promptSRC~\cite{khattak2023self}, $\lambda$ is set to 25.0. We utilize the Adam optimizer with a learning rate of 0.0025. For training, we use a batch size of 128 on the large-scale ImageNet dataset and a batch size of 32 for the other ten datasets. The model is trained for 50 epochs for base-to-novel generalization and few-shot classification tasks. For cross-dataset evaluation, the maximum number of epochs is set to 3. Evaluation metrics include top-1 accuracy and the harmonic mean (HM) between base and novel class accuracies to assess generalization. All experiments are conducted on an NVIDIA RTX 3090 GPU. 

\subsection{Base-to-Novel Generalization}
Table \ref{tab:base-to-new} presents the performance of InPK across 11 recognition datasets in the base-to-novel generalization setting. We implement all methods using a few-shot training approach, with 16 randomly sampled shots per base class. Compared to the recently proposed method TCP~\cite{yao2024tcp}, our approach shows enhanced performance on 10 out of 11 datasets in terms of harmonic mean (HM), with only a slight drop in performance on the Food101 dataset relative to method TCP. For the average result, since we infuse class-specific prior knowledge into learnable tokens in Pki, these tokens are expected to learn the common features shared across modalities. This allows our method to perform better on novel class, improving the average accuracy of novel class by 3.43\%. Additionally, InPK maintains strong performance on the base class as it can infer more discriminative text features, with an increase of 1.04\% in accuracy. When considering both base and novel classes, InPK achieves an absolute average gain of 2.34\% over TCP in terms of HM, demonstrating an effective balance between in-domain and out-of-domain data performance.

Table \ref{tab:other_method} presents a comparison of the average performance of our method against more recent prompt-tuning methods across 11 datasets. The results evident that our method outperforms other methods on both base and novel classes, which demonstrates that the InPK not only maintains excellent recognition accuracy on base class but also exhibits strong generalization capabilities for novel class. Moreover, HPT~\cite{wang2024learning} and CoPrompt~\cite{royconsistency} are the recently proposed methods that introduce prior knowledge. It can be seen that the performance of these methods that introduce prior knowledge but are not related to learnable tokens is lower than that of our proposed method, which verifies the importance of infusing prior knowledge into learnable tokens. 

\subsection{Few-Shot Learning}
We further investigate the robustness of our InPK in the few-shot learning setting. Figure \ref{fig:few-shot} presents the performance of InPK across different shots (1, 2, 4, 8, and 16 shots per class) on 11 datasets. We compare InPK with recent prompt tuning methods, including CoOp, CoCoOp, and TCP, achieving the best results on the majority of datasets. Compared to the previous state-of-the-art method TCP, InPK achieves an average improvement of 1.74\% across different shot settings. Notably, the performance gain becomes more pronounced as the number of shots decreases. This is attributed to the incorporation of class-specific prior knowledge, which serves as additional informative cues to compensate for the scarcity of labeled samples, thereby enhancing the model's ability to perceive fine-grained differences. This demonstrates that InPK maintains high performance even in limited data scenarios.

\subsection{Cross-Dataset Evaluation}
To further verify the generalization ability of the model, we evaluate the model and state-of-the-art methods in a cross-data setting. As shown in Table \ref{tab:cross-data}, our model exhibits outstanding generalization on 8 out of the 10 datasets, with a significant improvement over other methods. The average accuracy reaches 69.90\%, demonstrating strong robustness across varying data distributions. This is because the model infuses the prior knowledge into learnable tokens at various feature levels, which guides the model to focus on category information and better discover latent representations of novel class attributes, thereby improving the model's performance on unseen classes.

\begin{table*}[!tb]
    \small \centering
    \renewcommand{\arraystretch}{1.2}
 \setlength{\tabcolsep}{8pt}
  \caption{ Comparison of InPK with existing methods in the domain generalization evaluation setting.} 
    \scalebox{0.9}[0.9]{
    \begin{tabular}{l c@{\hspace{20pt}}ccccc}
    \toprule
    &  \textbf{{Source}} & \multicolumn{5}{c}{\textbf{ {Target}}} \\ \cmidrule(lr){2-2} \cmidrule(lr){3-7}
     & \textbf{ImageNet} & \textbf{ImageNetV2} & \textbf{ImageNet-S} & \textbf{ImageNet-A} & \textbf{ImageNet-R}  & \textbf{Average}\\
    \midrule 
    CoOp & 71.51 & 49.71 & 75.21 & 47.99 & 64.20  & 59.28  \\
    CoCoOp & 71.02 & 50.63 & 76.18 & 48.75 & 64.07 & 59.91  \\
     CoPrompt & 70.80  & 50.50  &\textbf{{77.51}} & 49.43 & 64.25 & 60.42  \\
    MMA   & 71.00    & 51.12 & 77.32 & 49.13 & 64.33 & 60.48  \\
    TCP & 71.20  & \textbf{{51.20}} & 76.73 & \textbf{{49.50}} & 64.60  & 60.51  \\
             \midrule
  \textbf{Ours}  & \textbf{71.77} & 51.07 & 77.47 & 49.27 &\textbf{{64.63}} & \textbf{60.61}  \\
                \bottomrule
    \end{tabular}
}
    \label{tab:robustness}
   \vspace{8pt}
\end{table*}

\subsection{Domain Generalization}
To evaluate the model's robustness on out-of-distribution datasets, we conduct a domain generalization evaluation. Table \ref{tab:robustness} compares our method with existing approaches in this setting. InPK demonstrates competitive performance, achieving the highest average accuracy across the four ImageNet variant datasets.

\subsection{Ablation Study} 

\begin{figure}[]
  \centering
 
     \scalebox{1}{
        \includegraphics[width=1\linewidth]{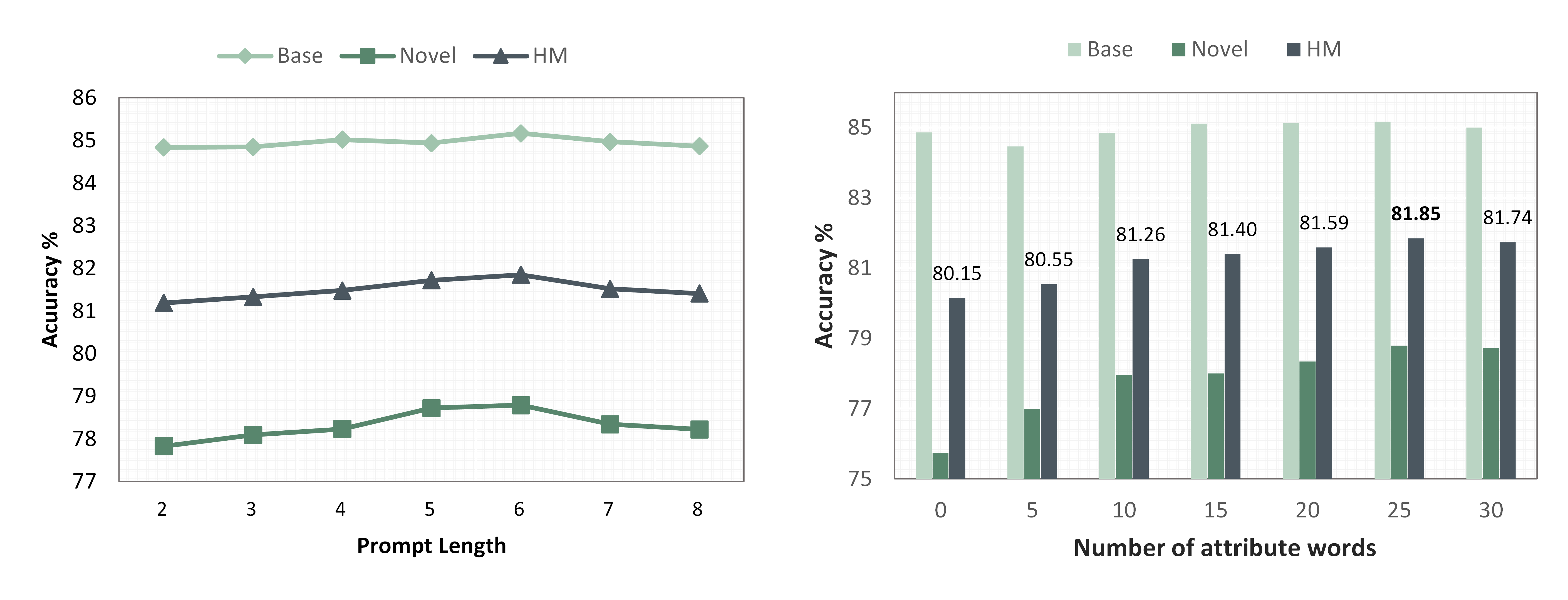}
     
     }

   \vspace{-5pt}
   \caption{Ablation on prompt depth (left) and the number of attribute words (right).}
   \label{fig:words}
    \vspace{-15pt}
\end{figure}

\noindent \textbf{Impact of prompt length.}
We study the effect of prompt length \( M \) on model performance in a base-to-novel generalization setting. As shown in Figure \ref{fig:words} (left), changing \( M \) has minimal impact on base class performance. However, for the novel class, performance improves as the prompt length increases. Ultimately, we select \( M=6 \), which yields the highest harmonic mean (HM) accuracy, as the optimal prompt length for the model.

\noindent \textbf{Impact of the number of attribute words.} We analyze the impact of the number of attribute words \( N \) on model performance in the base-to-novel generalization setting as shown in Figure \ref{fig:words} (right). By varying the number of attribute words, we observe the accuracy of the base class, new class, and harmonic mean (HM).  Overall, as the number of attribute words increases, the HM index gradually improves. This is because richer class-specific textual knowledge provides more comprehensive supervision for the model. Notably, when the number of attribute words \( N=25 \), the HM value is the highest, indicating that this amount of detailed prior knowledge optimizes the balance between base and novel class performance.

\noindent \textbf{Impact of inserting attribute-aware attention into different Transformer layers.} 
Table \ref{tab:layer} presents the analysis of inserting attribute-aware attention into different Transformer layers within PKi in the base-to-new generalization setting. The results indicate that even with $J=1$, where attribute-aware attention is only applied in the initialization stage, the model’s performance shows a significant improvement. This highlights the importance of providing a prior knowledge-based initialization for the learnable tokens. The model performance is observed to improve with the increase of the prompt depth of module insertion, which can be attributed to the fact that class-specific prior knowledge is able to reduce the loss of category information. Additionally, inserting attribute-aware attention beyond the ninth layer decreases performance, likely because earlier layers have already captured high-level features. Further insertions may interfere with these representations, making the model overly sensitive and reducing its generalization ability. The optimal result is obtained by incorporating attribute-aware attention into the transformer layers up to a depth \( J = 9 \).
\begin{table}[h]
\caption{ Impact of insert layers.} 
    \small \centering
    \renewcommand{\arraystretch}{1.2}
    \scalebox{1}[1]{
    \begin{tabular}{l|c|c|c|c|c}
    \toprule
    \textbf{Layer}& $\boldsymbol{\left\{1\right\}}$& $\boldsymbol{\left\{1\textendash3\right\}}$ &  $\boldsymbol{\left\{1 \textendash 6\right\}}$ &  $\boldsymbol{\left\{1 \textendash 9\right\}}$&  $\boldsymbol{\left\{1 \textendash 12\right\}}$\\
    \midrule 
    Base &   {84.91} &85.10 &85.06 & \textbf{85.17}&85.02 \\
    Novel &   {78.21} & 78.39& 78.53& 78.79 &\textbf{78.82} \\
    HM &   {81.42} &81.61 & 81.66  & \textbf{81.85} & 81.80\\

             \bottomrule
    \end{tabular}
    }
       
    \label{tab:layer}
    \vspace{-5pt}
\end{table} 

\noindent \textbf{Impact of prompt initialization strategy.} We evaluate the impact of different prompt initialization strategies, employing templates ``a photo of a \(\{\}\)" and ``XXXX \(\{\}\)” respectively. To align with the handcrafted template ``a photo of a \(\{\}\)", we set the prompt length in InPK to 4, represented as ``XXXX \(\{\}\)”. As shown in Table \ref{tab:initialization}, the performance differences between the two initialization methods are minimal. This outcome arises because the learnable tokens are subsequently infused with prior knowledge, making the model less sensitive to the choice of initialization.

\begin{table}[h]
\caption{Comparison of different prompt initialization strategies in the context of base-to-novel generalization, with average results across 11 datasets.} 
\small
    \centering
    \renewcommand{\arraystretch}{1.2}
     \setlength{\tabcolsep}{10pt}
    \scalebox{1}[1]
    {\begin{tabular}{ccccc}

    \toprule

  {\textbf{Initialization}}& \textbf{Base} & {\textbf{Novel}} & \textbf{HM} \\ %
    \midrule
    
     a photo of a \(\{\}\) & 84.84  & 78.28  & 81.43   \\
      X X X X \(\{\}\) & 85.02  & 78.23  & 81.48   \\
    \bottomrule 
    \end{tabular}}
    
    \label{tab:initialization}
        \vspace{-5pt}
\end{table}

\noindent \textbf{Impact of prior knowledge infusion strategy.} We experimented with strategies for infusing prior knowledge into learnable tokens, including `Concat', `MLP', and `AAA'. In the `Concat' strategy, The prior knowledge and the learnable tokens are unrelated, they are concatenated and sent to the subsequent encoder layers. In the `MLP' strategy, the two are passed through a fully connected layer to produce updated learnable tokens. Lastly, in the `AAA' strategy, they are processed through the attribute-aware attention module to generate the new learnable tokens. Experimental results in Table \ref{tab:Infusion} show that the `AAA' strategy achieves the best performance. This is because `Concat' or' MLP' tends to incorporate irrelevant information directly, leading to a decline in the quality of the fused representation. In contrast, the attribute-aware attention module employs a weight distribution strategy to automatically filter out irrelevant or unimportant details, reducing the influence of redundancy and noise. This enables the model to selectively focus on the most valuable aspects of prior knowledge, resulting in fused learnable tokens that are both more discriminative and generalizable.
\begin{table}[h]
\small
  \caption{Comparison of different prior knowledge infusion strategies for base-to-novel generalization across 11 datasets.} 
    \centering
    \renewcommand{\arraystretch}{1.2}
     \setlength{\tabcolsep}{10pt}
    \scalebox{1}[1]
    {\begin{tabular}{ccccc}

    \toprule

  {\textbf{Infusion Strategy}}& \textbf{Base} & {\textbf{Novel}} & \textbf{HM} \\ %
    \midrule
    
     Concat  & 83.63  & 77.95  & 80.69   \\
     MLP  & 85.02  & 78.23  & 81.48   \\
     AAA  & 85.17  & 78.79  & 81.85   \\
    \bottomrule 
    \end{tabular}}

    \label{tab:Infusion}
        \vspace{-10pt}
\end{table}

\noindent \textbf{Analysis of model components.} We conduct a comprehensive analysis of the key components of InPK, as illustrated in Table \ref{tab: component}. Starting from the baseline CoOp~\cite{zhou2022learning} (\textbf{\#1}), which fine-tunes VLMs with learnable tokens, we observe poor performance on novel classes due to overfitting to base classes. When prior knowledge is introduced to the model (\textbf{\#2}), we observe an improvement of 7.2\% of HM. This gain is attributed to the richer class-specific information provided by prior knowledge, offering more comprehensive supervision. Due to the overlap of common attribute words among different classes, class confusion often occurs. To address this, regularization \( L_{text}\) is introduced (\textbf{\#3}), further improving the model's performance by 1.08\%.  Building on this, we integrate the attribute-aware attention module (\textbf{\#4}), which infuses prior knowledge into the learnable tokens and facilitates their interaction with prior knowledge across multiple feature levels. This enhancement allows the model to explicitly focus on class-relevant information, resulting in a performance boost of 0.86\%. Finally, we propose a learnable text-to-vision projection layer to accommodate the adjustments in the text branch (\textbf{\#5}), specifically the introduction of prior knowledge. These additions ensure better visual and textual semantics alignment by simultaneously refining both branches. As a result, the model achieves superior performance on both base and novel classes, with the HM improving by 10.19\% compared to the baseline. In conclusion, the contribution of each model component to performance improvement is highlighted.
\begin{table}[h]
    \caption{Further analysis of the model components in the base-to-novel generalization setting across 11 datasets. Results are shown for base classes (B), novel classes (N), and harmonic mean (HM).}
\small
    \centering
    \renewcommand{\arraystretch}{1.2}
     \setlength{\tabcolsep}{7pt}
    \scalebox{1}[1]
    {\begin{tabular}{lcccccc}
    \toprule
  {\textbf{Dataset}}&  &\textbf{\#1} & {\textbf{\#2}} & \textbf{\#3} & \textbf{\#4}& \textbf{\#5}\\ %
    \midrule
    
       &B &76.47  & 76.50  & 76.17  & 77.30  & 77.60  \\
     ImNet  &N &67.88 & 70.40  & 72.37  & 71.73  & 71.37  \\
       &HM &71.92 & 73.32  & 74.22  & 74.41  & 74.35 \\
     \midrule 
      & B &98.00  & 98.00  & 98.33  & 98.33  & 98.67  \\
     Caltech & N &89.81   & 94.43  & 95.13  & 95.33  & 94.30    \\
      & HM &93.73 & 96.18  & 96.70  & 96.81  & 96.44  \\
     \midrule 
      & B&93.67 & 94.40  & 95.07  & 94.47  & 94.67 \\
     Pets & N&95.29 & 94.57  & 97.50  & 97.63  & 97.63 \\
      & HM&94.47  & 94.57  & 97.50  & 97.63  & 97.63  \\
     \midrule 
      & B&78.12 & 78.07  & 77.50  & 80.57  & 81.97  \\
     Cars & N&60.40  & 73.00  & 74.67  & 75.17  & 76.10  \\ 
      & HM&68.13 & 75.45  & 76.06  & 77.78  & 78.93 \\
     \midrule 
      & B&97.60 & 97.53  & 97.50  & 98.43  & 98.17 \\
     Flowers & N&59.67 & 77.80  & 82.50  & 80.27  & 81.27  \\
      & HM&74.06  & 86.55  & 89.38  & 88.43  & 88.92  \\
     \midrule 
      & B&88.33  & 89.50  & 90.63  & 90.07  & 90.10  \\
     Food &N &82.26 & 81.53  & 81.00  & 80.67  & 81.13  \\
      & HM&85.19   & 89.73  & 90.91  & 90.81  & 90.83  \\
     \midrule 
      & B&40.44 & 40.17  & 38.83  & 43.50  & 45.67  \\
     Aircraft & N&22.30& 35.40  & 38.63  & 37.40  & 36.50  \\
      &HM &28.75 & 37.63  & 38.73  & 40.22  & 40.57  \\
     \midrule 
      &B &80.60  & 80.60  & 80.27  & 82.57  & 82.77  \\
     SUN397 &N &65.89  & 81.53  & 81.00  & 80.67  & 81.13  \\
      &HM &72.51 & 81.06  & 80.63  & 81.61  & 81.94  \\
     \midrule 
      &B &79.44 & 81.10  & 80.87  & 82.60  & 84.43  \\
    DTD &N &41.18 & 67.07  & 69.30  & 69.83  & 71.30  \\
     &HM &54.24 & 73.42  & 74.64  & 75.68  & 77.31  \\
    \midrule 
      & B&92.19 & 91.10  & 90.47  & 91.63  & 95.70  \\
     EuroSAT &N &54.74   & 63.23  & 64.93  & 71.77  & 82.37  \\
      &HM &68.69 & 74.65  & 75.60  & 80.49  & 88.54  \\
     \midrule 
      &B &84.69 & 85.23  & 85.03  & 86.37  & 87.13  \\
     UCF & N&56.05& 79.50  & 82.93  & 83.30  & 83.10  \\
      &HM &67.46  & 82.27  & 83.97  & 84.81  & 85.07  \\
     \midrule 
      &B &82.69  & 82.93  & 82.79  & 84.17  & \textbf{85.17} \\
     Average & N&63.22& 75.17  & 77.29  & 77.70  & \textbf{78.79}  \\
      &HM &71.66  & 78.86  & 79.94  & 80.80  & \textbf{81.85} \\
    \bottomrule 
    \end{tabular}}

    \label{tab: component}
\end{table}

To further verify the contributions of both prior knowledge and attribute-aware attention to model performance, we conduct a comprehensive analysis of the key components in the text branch, as illustrated in Table \ref{tab:ablation}. In this analysis, the learnable projection layer from text prompts to visual prompts is consistently retained. In this analysis, \textit{Know.} represents the input attribute words, \textit{Attn.} refers to the attribute-aware attention. In the text branch, when only class-specific prior knowledge is applied, the model exhibits improvement with an average gain from 79.03\% to 80.69\% due to the additional supervision provided by the rich attribute information. On the other hand, when only attribute-aware attention is used, the model strikes a better balance between base and novel classes with an average improvement of 0.92\%. When both components (\textit{Know.} and \textit{Attn.}) are incorporated together, the model achieves substantial gains across both base and new classes (83.70\% \textit{vs} 85.17\%, 74.86\% \textit{vs} 78.79\%). These results demonstrate that model performance is not solely influenced by prior knowledge itself but also by the manner in which the prior knowledge is processed. Specifically, the combination of prior knowledge with learnable tokens through the attribute-aware attention mechanism plays a crucial role in enhancing the model's performance.

\begin{table}[h]
 \caption{Ablation on attribute words, attribute-aware attention and learnable text-to-vision projection layer in InPK.}   
    \small \centering
    \renewcommand{\arraystretch}{1}
 \setlength{\tabcolsep}{6pt}

    \scalebox{1}[1]{
    \begin{tabular}{ccccc}
    \toprule
    \multicolumn{2}{c}{\textbf {Text encoder}} &   \multirow{2.5}{*}{\textbf{Base}} &  \multirow{2.5}{*}{\textbf{Novel}} &  \multirow{2.5}{*}{\textbf{HM}}\\ \cmidrule(lr){1-2} 
     \textbf{\textit{Know.}} & \textbf{\textit{Attn.}}   \\
    \midrule 
     \ding{55} &  \ding{55}   & 83.70 &74.86 & 79.03  \\
    \ding{51} & \ding{55}& 83.63
 &77.95 & 80.69  \\
    \ding{55} &  \ding{51} &84.86 &75.94 & 80.15\\
   \ding{51} & \ding{51}  & \textbf{85.17} &\textbf{78.79} & \textbf{81.85}  \\

    \bottomrule
    \end{tabular}
    }
     
    \label{tab:ablation}
      \vspace{4pt}
\end{table} 

\noindent \textbf{Effectiveness of prior knowledge infusion.}
To further assess the impact of prior knowledge infusion (pkin) into learnable tokens, we apply pkin only before the first encoder layer in method CoOp~\cite{zhou2022learning}, KgCoOp~\cite{yao2023visual}, and PromptSRC~\cite{khattak2023self}. As shown in Table \ref{tab:robust initialization}, all three methods exhibit significant improvement, confirming the effectiveness of pkin.

\begin{table}[h]
 \caption{Comparison of baseline methods with and without pkin in the base-to-novel generalization setting. `pkin’ denotes prior knowledge infusion into learnable tokens before feeding them into the encoder layer.} 
\small
    \centering
     \setlength{\tabcolsep}{10pt}
    \scalebox{1}[1]
    {\begin{tabular}{lcccc}

    \toprule

  {\textbf{Method}}& \textbf{Base} & {\textbf{Novel}} & \textbf{HM} \\ %
    \midrule
    
     CoOp  & 82.69  & 63.22  & 71.66   \\
    + \textbf{pkin} & \textbf{82.71\textcolor{red}{\tiny{+0.02}}}  & \textbf{71.42\textcolor{red}{\tiny{+8.20}}}  & \textbf{76.65\textcolor{red}{\tiny{+4.99}}} \\
    \midrule
      KgCoOp  & 80.73 & 73.60 &  77.00 \\
    + \textbf{pkin}  & \textbf{84.16\textcolor{red}{\tiny{+3.43}}}  & \textbf{77.93\textcolor{red}{\tiny{+4.33}}}  & \textbf{80.93\textcolor{red}{\tiny{+3.93}}}  \\
    \midrule
      PromptSRC  & 84.26  & 76.10  & 79.97   \\
    + \textbf{pkin}  & \textbf{85.15\textcolor{red}{\tiny{+0.89}}}  & \textbf{78.60\textcolor{red}{\tiny{+2.50}}}  & \textbf{81.74\textcolor{red}{\tiny{+1.77}}} \\
    \bottomrule 
    \end{tabular}}
   
    \label{tab:robust initialization}
\end{table}

\section{Conclusion} 
In this paper, we have proposed InPK to tackle the issue of prompt-tuning methods overfitting to seen classes and experiencing domain shifts on unseen ones. We infuse class-specific prior knowledge into the learnable tokens at the initialization stage and progressively reinforce their interaction with prior knowledge across multiple feature levels. InPK guides the model to focus on class-relevant information, allowing learnable tokens to capture the fine-grained differences and universal visual concepts within prior knowledge more effectively, thereby enabling the model to extract more discriminative and generalized text features. Furthermore, we introduce a learnable text-to-vision projection layer to accommodate text adjustments, enhancing visual-text alignment. Our method has demonstrated strong performance across 11 recognition datasets, outperforming state-of-the-art approaches in base-to-novel generalization, few-shot learning, and cross-dataset evaluation tasks.

\section{Acknowledge} 
This work was supported in part by the National Natural Science Foundation of China under Grant 62306067, and Grant 62220106008, in part by Sichuan Science and Technology Program under Grant 2024NSFSC1463, in part by Guangdong Basic and Applied Basic Research Foundation under grant No. 2025A1515010108.

\bibliographystyle{IEEEtran}
\bibliography{reference}

\begin{thebibliography}{10}
\providecommand{\url}[1]{#1}
\csname url@samestyle\endcsname
\providecommand{\newblock}{\relax}
\providecommand{\bibinfo}[2]{#2}
\providecommand{\BIBentrySTDinterwordspacing}{\spaceskip=0pt\relax}
\providecommand{\BIBentryALTinterwordstretchfactor}{4}
\providecommand{\BIBentryALTinterwordspacing}{\spaceskip=\fontdimen2\font plus
\BIBentryALTinterwordstretchfactor\fontdimen3\font minus \fontdimen4\font\relax}
\providecommand{\BIBforeignlanguage}[2]{{%
\expandafter\ifx\csname l@#1\endcsname\relax
\typeout{** WARNING: IEEEtran.bst: No hyphenation pattern has been}%
\typeout{** loaded for the language `#1'. Using the pattern for}%
\typeout{** the default language instead.}%
\else
\language=\csname l@#1\endcsname
\fi
#2}}
\providecommand{\BIBdecl}{\relax}
\BIBdecl

\bibitem{radford2021learning}
A.~Radford, J.~W. Kim, C.~Hallacy, A.~Ramesh, G.~Goh, S.~Agarwal, G.~Sastry, A.~Askell, P.~Mishkin, J.~Clark \emph{et~al.}, ``Learning transferable visual models from natural language supervision,'' in \emph{International conference on machine learning}.\hskip 1em plus 0.5em minus 0.4em\relax PmLR, 2021, pp. 8748--8763.

\bibitem{gao2024clip}
P.~Gao, S.~Geng, R.~Zhang, T.~Ma, R.~Fang, Y.~Zhang, H.~Li, and Y.~Qiao, ``Clip-adapter: Better vision-language models with feature adapters,'' \emph{International Journal of Computer Vision}, vol. 132, no.~2, pp. 581--595, 2024.

\bibitem{zhang2022tip}
R.~Zhang, W.~Zhang, R.~Fang, P.~Gao, K.~Li, J.~Dai, Y.~Qiao, and H.~Li, ``Tip-adapter: Training-free adaption of clip for few-shot classification,'' in \emph{European conference on computer vision}.\hskip 1em plus 0.5em minus 0.4em\relax Springer, 2022, pp. 493--510.

\bibitem{zhou2022learning}
K.~Zhou, J.~Yang, C.~C. Loy, and Z.~Liu, ``Learning to prompt for vision-language models,'' \emph{International Journal of Computer Vision}, vol. 130, no.~9, pp. 2337--2348, 2022.

\bibitem{zhou2022conditional}
{K. Zhou, J. Yang, C. C. Loy, and Z. Liu}, ``Conditional prompt learning for vision-language models,'' in \emph{Proceedings of the IEEE/CVF conference on computer vision and pattern recognition}, 2022, pp. 16\,816--16\,825.

\bibitem{khattak2023maple}
M.~U. Khattak, H.~Rasheed, M.~Maaz, S.~Khan, and F.~S. Khan, ``Maple: Multi-modal prompt learning,'' in \emph{Proceedings of the IEEE/CVF conference on computer vision and pattern recognition}, 2023, pp. 19\,113--19\,122.

\bibitem{khattak2023self}
M.~U. Khattak, S.~T. Wasim, M.~Naseer, S.~Khan, M.-H. Yang, and F.~S. Khan, ``Self-regulating prompts: Foundational model adaptation without forgetting,'' in \emph{Proceedings of the IEEE/CVF international conference on computer vision}, 2023, pp. 15\,190--15\,200.

\bibitem{jia2022visual}
M.~Jia, L.~Tang, B.-C. Chen, C.~Cardie, S.~Belongie, B.~Hariharan, and S.-N. Lim, ``Visual prompt tuning,'' in \emph{European conference on computer vision}.\hskip 1em plus 0.5em minus 0.4em\relax Springer, 2022, pp. 709--727.

\bibitem{xing2023dual}
Y.~Xing, Q.~Wu, D.~Cheng, S.~Zhang, G.~Liang, P.~Wang, and Y.~Zhang, ``Dual modality prompt tuning for vision-language pre-trained model,'' \emph{IEEE Transactions on Multimedia}, vol.~26, pp. 2056--2068, 2023.

\bibitem{lee2024coapt}
G.~Lee, S.~An, S.~Baik, and S.~Lee, ``Coapt: Context attribute words for prompt tuning,'' \emph{arXiv preprint arXiv:2407.13808}, 2024.

\bibitem{menon2022visual}
S.~Menon and C.~Vondrick, ``Visual classification via description from large language models,'' \emph{arXiv preprint arXiv:2210.07183}, 2022.

\bibitem{wang2024learning}
Y.~Wang, X.~Jiang, D.~Cheng, D.~Li, and C.~Zhao, ``Learning hierarchical prompt with structured linguistic knowledge for vision-language models,'' in \emph{Proceedings of the AAAI conference on artificial intelligence}, vol.~38, no.~6, 2024, pp. 5749--5757.

\bibitem{kan2023knowledge}
B.~Kan, T.~Wang, W.~Lu, X.~Zhen, W.~Guan, and F.~Zheng, ``Knowledge-aware prompt tuning for generalizable vision-language models,'' in \emph{Proceedings of the IEEE/CVF conference on computer vision and pattern recognition}, 2023, pp. 15\,670--15\,680.

\bibitem{royconsistency}
S.~Roy and A.~Etemad, ``Consistency-guided prompt learning for vision-language models,'' in \emph{The Twelfth International Conference on Learning Representations}.

\bibitem{zhang2024concept}
Y.~Zhang, C.~Zhang, K.~Yu, Y.~Tang, and Z.~He, ``Concept-guided prompt learning for generalization in vision-language models,'' in \emph{Proceedings of the AAAI conference on artificial intelligence}, vol.~38, no.~7, 2024, pp. 7377--7386.

\bibitem{wei2021meta}
J.~Wei, X.~Xu, Z.~Wang, and G.~Wang, ``Meta self-paced learning for cross-modal matching,'' in \emph{Proceedings of the 29th ACM international conference on multimedia}, 2021, pp. 3835--3843.

\bibitem{9454290}
J.~Wei, Y.~Yang, X.~Xu, X.~Zhu, and H.~T. Shen, ``Universal weighting metric learning for cross-modal retrieval,'' \emph{IEEE transactions on pattern analysis and machine intelligence}, vol.~44, no.~10, pp. 6534--6545, 2022.

\bibitem{wei2020universal}
J.~Wei, X.~Xu, Y.~Yang, Y.~Ji, Z.~Wang, and H.~T. Shen, ``Universal weighting metric learning for cross-modal matching,'' in \emph{Proceedings of the IEEE/CVF conference on computer vision and pattern recognition}, 2020, pp. 13\,005--13\,014.

\bibitem{jia2021scaling}
C.~Jia, Y.~Yang, Y.~Xia, Y.-T. Chen, Z.~Parekh, H.~Pham, Q.~Le, Y.-H. Sung, Z.~Li, and T.~Duerig, ``Scaling up visual and vision-language representation learning with noisy text supervision,'' in \emph{International conference on machine learning}.\hskip 1em plus 0.5em minus 0.4em\relax PMLR, 2021, pp. 4904--4916.

\bibitem{yao2021filip}
L.~Yao, R.~Huang, L.~Hou, G.~Lu, M.~Niu, H.~Xu, X.~Liang, Z.~Li, X.~Jiang, and C.~Xu, ``Filip: Fine-grained interactive language-image pre-training,'' \emph{arXiv preprint arXiv:2111.07783}, 2021.

\bibitem{yuan2021florence}
L.~Yuan, D.~Chen, Y.-L. Chen, N.~Codella, X.~Dai, J.~Gao, H.~Hu, X.~Huang, B.~Li, C.~Li \emph{et~al.}, ``Florence: A new foundation model for computer vision,'' \emph{arXiv preprint arXiv:2111.11432}, 2021.

\bibitem{liu2023learning}
H.~Liu, K.~Son, J.~Yang, C.~Liu, J.~Gao, Y.~J. Lee, and C.~Li, ``Learning customized visual models with retrieval-augmented knowledge,'' in \emph{Proceedings of the IEEE/CVF conference on computer vision and pattern recognition}, 2023, pp. 15\,148--15\,158.

\bibitem{wysoczanska2024clip}
M.~Wysocza{\'n}ska, M.~Ramamonjisoa, T.~Trzci{\'n}ski, and O.~Sim{\'e}oni, ``Clip-diy: Clip dense inference yields open-vocabulary semantic segmentation for-free,'' in \emph{Proceedings of the IEEE/CVF conference on computer vision and pattern recognition}, 2024, pp. 1403--1413.

\bibitem{zhang2024exploring}
Y.~Zhang, M.-H. Guo, M.~Wang, and S.-M. Hu, ``Exploring regional clues in clip for zero-shot semantic segmentation,'' in \emph{Proceedings of the IEEE/CVF conference on computer vision and pattern recognition}, 2024, pp. 3270--3280.

\bibitem{zhou2022extract}
C.~Zhou, C.~C. Loy, and B.~Dai, ``Extract free dense labels from clip,'' in \emph{European conference on computer vision}.\hskip 1em plus 0.5em minus 0.4em\relax Springer, 2022, pp. 696--712.

\bibitem{wang2024cm}
W.~Wang, X.~He, Y.~Zhang, L.~Guo, J.~Shen, J.~Li, and J.~Liu, ``Cm-masksd: Cross-modality masked self-distillation for referring image segmentation,'' \emph{IEEE Transactions on Multimedia}, vol.~26, pp. 6906--6916, 2024.

\bibitem{zhang2021tip}
R.~Zhang, R.~Fang, W.~Zhang, P.~Gao, K.~Li, J.~Dai, Y.~Qiao, and H.~Li, ``Tip-adapter: Training-free clip-adapter for better vision-language modeling,'' \emph{arXiv preprint arXiv:2111.03930}, 2021.

\bibitem{alayrac2022flamingo}
J.-B. Alayrac, J.~Donahue, P.~Luc, A.~Miech, I.~Barr, Y.~Hasson, K.~Lenc, A.~Mensch, K.~Millican, M.~Reynolds \emph{et~al.}, ``Flamingo: a visual language model for few-shot learning,'' \emph{Advances in neural information processing systems}, vol.~35, pp. 23\,716--23\,736, 2022.

\bibitem{yao2023visual}
H.~Yao, R.~Zhang, and C.~Xu, ``Visual-language prompt tuning with knowledge-guided context optimization,'' in \emph{Proceedings of the IEEE/CVF conference on computer vision and pattern recognition}, 2023, pp. 6757--6767.

\bibitem{yang2024mma}
L.~Yang, R.-Y. Zhang, Y.~Wang, and X.~Xie, ``Mma: Multi-modal adapter for vision-language models,'' in \emph{Proceedings of the IEEE/CVF conference on computer vision and pattern recognition}, 2024, pp. 23\,826--23\,837.

\bibitem{wei2024runge}
J.~Wei, Y.~Yang, X.~Guan, X.~Xu, G.~Wang, and H.~T. Shen, ``Runge-kutta guided feature augmentation for few-sample learning,'' \emph{IEEE Transactions on Multimedia}, 2024.

\bibitem{wu2024fine}
Q.~Wu, J.~Qi, D.~Zhang, H.~Zhang, and J.~Tang, ``Fine-tuning for few-shot image classification by multimodal prototype regularization,'' \emph{IEEE Transactions on Multimedia}, 2024.

\bibitem{fang2024simple}
R.~Fang, G.~Pang, and X.~Bai, ``Simple image-level classification improves open-vocabulary object detection,'' in \emph{Proceedings of the AAAI Conference on Artificial Intelligence}, vol.~38, no.~2, 2024, pp. 1716--1725.

\bibitem{zang2022open}
Y.~Zang, W.~Li, K.~Zhou, C.~Huang, and C.~C. Loy, ``Open-vocabulary detr with conditional matching,'' in \emph{European conference on computer vision}.\hskip 1em plus 0.5em minus 0.4em\relax Springer, 2022, pp. 106--122.

\bibitem{zhou2022detecting}
X.~Zhou, R.~Girdhar, A.~Joulin, P.~Kr{\"a}henb{\"u}hl, and I.~Misra, ``Detecting twenty-thousand classes using image-level supervision,'' in \emph{European conference on computer vision}.\hskip 1em plus 0.5em minus 0.4em\relax Springer, 2022, pp. 350--368.

\bibitem{zhong2022regionclip}
Y.~Zhong, J.~Yang, P.~Zhang, C.~Li, N.~Codella, L.~H. Li, L.~Zhou, X.~Dai, L.~Yuan, Y.~Li \emph{et~al.}, ``Regionclip: Region-based language-image pretraining,'' in \emph{Proceedings of the IEEE/CVF conference on computer vision and pattern recognition}, 2022.

\bibitem{ke2024vldadaptor}
J.~Ke, L.~He, B.~Han, J.~Li, D.~Wang, and X.~Gao, ``Vldadaptor: Domain adaptive object detection with vision-language model distillation,'' \emph{IEEE Transactions on Multimedia}, 2024.

\bibitem{liu2024multi}
X.~Liu, J.~Wu, W.~Yang, X.~Zhou, and T.~Zhang, ``Multi-modal attribute prompting for vision-language models,'' \emph{IEEE Transactions on Circuits and Systems for Video Technology}, 2024.

\bibitem{cho2023distribution}
E.~Cho, J.~Kim, and H.~J. Kim, ``Distribution-aware prompt tuning for vision-language models,'' in \emph{Proceedings of the IEEE/CVF international conference on computer vision}, 2023, pp. 22\,004--22\,013.

\bibitem{zhang2024unleash}
W.~Zhang, L.~Wu, Z.~Zhang, T.~Yu, C.~Ma, X.~Jin, X.~Yang, and W.~Zeng, ``Unleash the power of vision-language models by visual attention prompt and multi-modal interaction,'' \emph{IEEE Transactions on Multimedia}, 2024.

\bibitem{yao2024tcp}
H.~Yao, R.~Zhang, and C.~Xu, ``Tcp: Textual-based class-aware prompt tuning for visual-language model,'' in \emph{Proceedings of the IEEE/CVF conference on computer vision and pattern recognition}, 2024, pp. 23\,438--23\,448.

\bibitem{zhu2023prompt}
B.~Zhu, Y.~Niu, Y.~Han, Y.~Wu, and H.~Zhang, ``Prompt-aligned gradient for prompt tuning,'' in \emph{Proceedings of the IEEE/CVF international conference on computer vision}, 2023, pp. 15\,659--15\,669.

\bibitem{li2023gradient}
J.~Li, M.~Gao, L.~Wei, S.~Tang, W.~Zhang, M.~Li, W.~Ji, Q.~Tian, T.-S. Chua, and Y.~Zhuang, ``Gradient-regulated meta-prompt learning for generalizable vision-language models,'' in \emph{Proceedings of the IEEE/CVF international conference on computer vision}, 2023, pp. 2551--2562.

\bibitem{park2024prompt}
J.~Park, J.~Ko, and H.~J. Kim, ``Prompt learning via meta-regularization,'' in \emph{Proceedings of the IEEE/CVF conference on computer vision and pattern recognition}, 2024, pp. 26\,940--26\,950.

\bibitem{vaswani2017attention}
A.~Vaswani, N.~Shazeer, N.~Parmar, J.~Uszkoreit, L.~Jones, A.~N. Gomez, {\L}.~Kaiser, and I.~Polosukhin, ``Attention is all you need,'' \emph{Advances in neural information processing systems}, vol.~30, 2017.

\bibitem{xian2017zero}
Y.~Xian, B.~Schiele, and Z.~Akata, ``Zero-shot learning-the good, the bad and the ugly,'' in \emph{Proceedings of the IEEE/CVF conference on computer vision and pattern recognition}, 2017, pp. 4582--4591.

\bibitem{deng2009imagenet}
J.~Deng, W.~Dong, R.~Socher, L.-J. Li, K.~Li, and L.~Fei-Fei, ``Imagenet: A large-scale hierarchical image database,'' in \emph{Proceedings of the IEEE/CVF conference on computer vision and pattern recognition}.\hskip 1em plus 0.5em minus 0.4em\relax Ieee, 2009, pp. 248--255.

\bibitem{fei2004learning}
L.~Fei-Fei, R.~Fergus, and P.~Perona, ``Learning generative visual models from few training examples: An incremental bayesian approach tested on 101 object categories,'' in \emph{2004 conference on computer vision and pattern recognition workshop}.\hskip 1em plus 0.5em minus 0.4em\relax IEEE, 2004, pp. 178--178.

\bibitem{parkhi2012cats}
O.~M. Parkhi, A.~Vedaldi, A.~Zisserman, and C.~Jawahar, ``Cats and dogs,'' in \emph{Proceedings of the IEEE/CVF conference on computer vision and pattern recognition}.\hskip 1em plus 0.5em minus 0.4em\relax IEEE, 2012, pp. 3498--3505.

\bibitem{krause20133d}
J.~Krause, M.~Stark, J.~Deng, and L.~Fei-Fei, ``3d object representations for fine-grained categorization,'' in \emph{Proceedings of the IEEE international conference on computer vision workshops}, 2013, pp. 554--561.

\bibitem{nilsback2008automated}
M.-E. Nilsback and A.~Zisserman, ``Automated flower classification over a large number of classes,'' in \emph{2008 Sixth Indian conference on computer vision, graphics \& image processing}.\hskip 1em plus 0.5em minus 0.4em\relax IEEE, 2008, pp. 722--729.

\bibitem{bossard2014food}
L.~Bossard, M.~Guillaumin, and L.~Van~Gool, ``Food-101--mining discriminative components with random forests,'' in \emph{Computer vision--ECCV 2014: 13th European conference, zurich, Switzerland, September 6-12, 2014, proceedings, part VI 13}.\hskip 1em plus 0.5em minus 0.4em\relax Springer, 2014, pp. 446--461.

\bibitem{maji2013fine}
S.~Maji, E.~Rahtu, J.~Kannala, M.~Blaschko, and A.~Vedaldi, ``Fine-grained visual classification of aircraft,'' \emph{arXiv preprint arXiv:1306.5151}, 2013.

\bibitem{xiao2010sun}
J.~Xiao, J.~Hays, K.~A. Ehinger, A.~Oliva, and A.~Torralba, ``Sun database: Large-scale scene recognition from abbey to zoo,'' in \emph{Proceedings of the IEEE/CVF conference on computer vision and pattern recognition}.\hskip 1em plus 0.5em minus 0.4em\relax IEEE, 2010, pp. 3485--3492.

\bibitem{soomro2012dataset}
K.~Soomro, A.~R. Zamir, and M.~Shah, ``A dataset of 101 human action classes from videos in the wild,'' \emph{Center for Research in Computer Vision}, vol.~2, no.~11, pp. 1--7, 2012.

\bibitem{cimpoi2014describing}
M.~Cimpoi, S.~Maji, I.~Kokkinos, S.~Mohamed, and A.~Vedaldi, ``Describing textures in the wild,'' in \emph{Proceedings of the IEEE/CVF conference on computer vision and pattern recognition}, 2014, pp. 3606--3613.

\bibitem{helber2019eurosat}
P.~Helber, B.~Bischke, A.~Dengel, and D.~Borth, ``Eurosat: A novel dataset and deep learning benchmark for land use and land cover classification,'' \emph{IEEE Journal of Selected Topics in Applied Earth Observations and Remote Sensing}, vol.~12, no.~7, pp. 2217--2226, 2019.

\bibitem{paszke2017automatic}
A.~Paszke, S.~Gross, S.~Chintala, G.~Chanan, E.~Yang, Z.~DeVito, Z.~Lin, A.~Desmaison, L.~Antiga, and A.~Lerer, ``Automatic differentiation in pytorch,'' 2017.

\bibitem{zhou2022domain}
K.~Zhou, Z.~Liu, Y.~Qiao, T.~Xiang, and C.~C. Loy, ``Domain generalization: A survey,'' \emph{IEEE transactions on pattern analysis and machine intelligence}, vol.~45, no.~4, pp. 4396--4415, 2022.

\end{thebibliography}

\vspace{-1em}

\begin{IEEEbiographynophoto}{Shuchang Zhou} is currently pursuing a Ph.D with the Center for Future Media of the University of Electronic Science and Technology of China (UESTC), under the supervision of Prof. Yang Yang. Her main research interests include computer vision, multimedia, and few-shot learning.

\end{IEEEbiographynophoto}

\vspace{-2em}

\begin{IEEEbiographynophoto}{Jiwei Wei} received his Ph.D. degree in June 2022 from the University of Electronic Science and Technology of China (UESTC), under the supervision of Prof. Yang Yang. In June 2022, Dr. Wei joined the School of Computer Science and Engineering, University of Electronic Science and Technology of China (UESTC) as a Postdoctoral Research Fellow. His current research interests include multimedia information retrieval, metric learning, and computer vision.

\end{IEEEbiographynophoto}

\vspace{-2em}

\begin{IEEEbiographynophoto}{Shiyuan He} is currently working toward the Ph.D degree with the School of Computer Science and Engineering, University of Electronic Science and Technology of China, Chengdu, China. His research interests include multimedia, computer vision, and machine learning.

\end{IEEEbiographynophoto}

\vspace{-2em}
\begin{IEEEbiographynophoto}{Yuyang Zhou} is an associate researcher in the School of Cyberspace Security, Hainan University, Haikou, P.R. China. She received the Ph.D. degree in the School of Computer Science and Engineering, University of Electronic
Science and Technology of China (UESTC), Chengdu, P.R. China in 2023.
Her research interests include applied cryptography and artificial intelligence
security

\end{IEEEbiographynophoto}

\vspace{-2em}
\begin{IEEEbiographynophoto}{Chaoning Zhang} received the B.S. and master’s degrees in electrical engineering from the Harbin Institute of Technology, Harbin, China, in 2012 and 2015, respectively, the master’s degree in engineering and policy analysis from TU Delft, Delft, The Netherlands, and the Ph.D. degree form the Robotics and Computer Vision Laboratory, Korea Advanced Institute of Science Technology, Daejeon, South Korea, in 2021. He is currently a Postdoctoral Researcher with RCV. His research interests include deep learning, adversarial machine learning, and deep hiding for steganography or watermarking.

\end{IEEEbiographynophoto}

\vspace{-2em}

\begin{IEEEbiographynophoto}{Jie Zou} received the doctorate degree from the University of Amsterdam, Amsterdam, the Netherlands. He was a Research Fellow with Nanyang Technological University, Singapore, and a Postdoc with the University of Amsterdam. He is currently a faculty with the School of Computer Science and Engineering, University of Electronic Science and Technology of China, Chengdu, China. He has authored or coauthored papers in ACM TOIS, SIGIR, CIKM, and IP\&M.His research interests include information retrieval, natural language processing, recommender systems, and multimedia.
\end{IEEEbiographynophoto}

\vspace{-2em}

\begin{IEEEbiographynophoto}{Ning Xie} received the M.E. and Ph.D. degrees from the Department of Computer Science, Tokyo Institute of Technology, Tokyo, Japan, in 2009 and 2012, respectively. In 2012, he was appointed as a Research Associate with Tokyo Institute of Technology. Since 2017, he has been an Associate Professor with the Center for Future Media, School of Computer Science and Engineering, University of Electronic Science and Technology of China (UESTC). His research interests include computer graphics, game engine, and the theory and application of artificial intelligence and machine learning. His research is supported by research grants, including NSFC, China, MOE, China, CREST, Japan, and The Ministry of Education, Culture, Sports, Science and Technology, Japan.

\end{IEEEbiographynophoto}

\vspace{-2em}
\begin{IEEEbiographynophoto}{Yang Yang} received the bachelor's and master's degrees from Jilin University and Peking University, in 2006 and 2009, respectively, the Ph.D. degree under the supervision of Prof. H. T. Shen and Prof. X. Zhou from The University of Queensland, Australia, in 2012. From 2012 to 2014, he was a Research Fellow under the supervision of Prof. T.-S. Chua with the National University of Singapore. He is currently with the University of Electronic Science and Technology of China.

\end{IEEEbiographynophoto}

\vfill

\end{document}